\documentclass[sigconf]{acmart}
\AtBeginDocument{%
  }

\copyrightyear{2026}
\acmYear{2026}
\setcopyright{cc}
\setcctype{by}
\acmConference[WWW '26]{Proceedings of the ACM Web Conference 2026}{April 13--17, 2026}{Dubai, United Arab Emirates}
\acmBooktitle{Proceedings of the ACM Web Conference 2026 (WWW '26), April 13--17, 2026, Dubai, United Arab Emirates}
\acmPrice{}
\acmDOI{10.1145/3774904.3792371}
\acmISBN{979-8-4007-2307-0/2026/04}

\usepackage{enumitem}
\usepackage{bm}
\usepackage{amsmath}
\usepackage{amsfonts}
\usepackage{subcaption}
\usepackage{multirow}
\usepackage{booktabs}
\usepackage[table]{xcolor}
\usepackage{colortbl}
\definecolor{lightblue}{HTML}{F0F8FF}

\begin{document}

\title{Adaptive Task Balancing for Visual Instruction Tuning via Inter-Task Contribution and Intra-Task Difficulty}


\author{Yanqi Dai}\authornote{Part of this work was done during Yanqi Dai’s internship at AMAP, Alibaba Group.}
\orcid{0009-0000-5441-3263}
\email{yanqidai@ruc.edu.cn}
\affiliation{
  \institution{Renmin University of China}
  \city{Beijing}
  \country{China}}

\author{Yong Wang}
\email{wangyong.lz@alibaba-inc.com}
\orcid{0009-0008-7878-493X}
\affiliation{
  \institution{AMAP, Alibaba Group}
  \city{Beijing}
  \country{China}}

\author{Zebin You}
\email{zebin@ruc.edu.cn}
\orcid{0009-0005-3577-8693}
\affiliation{
  \institution{Renmin University of China}
  \city{Beijing}
  \country{China}}

\author{Dong Jing}
\orcid{0009-0008-0769-8102}
\email{jingdong98@ruc.edu.cn}
\affiliation{
  \institution{Renmin University of China}
  \city{Beijing}
  \country{China}}

\author{Xiangxiang Chu}
\orcid{0000-0003-2548-0605}
\email{chuxiangxiang.cxx@alibaba-inc.com}
\affiliation{
  \institution{AMAP, Alibaba Group}
  \city{Beijing}
  \country{China}}

\author{Zhiwu Lu}\authornote{Corresponding author.}
\orcid{0000-0001-6429-7956}
\email{luzhiwu@ruc.edu.cn}
\affiliation{
  \institution{Renmin University of China}
  \city{Beijing}
  \country{China}}

\renewcommand{\shortauthors}{Yanqi Dai et al.}

\begin{abstract}
Visual instruction tuning is a key training stage of large multimodal models.
However, when learning multiple visual tasks simultaneously, this approach often results in suboptimal and imbalanced overall performance due to latent knowledge conflicts across tasks.
To mitigate this issue, we propose a novel \textbf{A}daptive \textbf{T}ask \textbf{B}alancing approach tailored for \textbf{vis}ual instruction tuning (\textbf{VisATB}).
Specifically, we measure two critical dimensions for visual task balancing based on validation performance: (1) \textit{Inter-Task Contribution}, the mechanism where learning one task enhances the performance on others owing to shared knowledge across tasks, and (2) \textit{Intra-Task Difficulty}, which denotes the inherent learning difficulty of a single task. 
Furthermore, we propose prioritizing three categories of tasks with greater weight: those that offer substantial contributions to others, those that receive minimal contributions from others, and those that present high learning difficulties.
Among these three task weighting strategies, the first and third focus on improving overall performance, and the second targets the mitigation of performance imbalance.
Extensive experiments on three benchmarks demonstrate that our VisATB approach consistently achieves superior and more balanced overall performance in visual instruction tuning.
The data, code, and models are available at \href{https://github.com/YanqiDai/VisATB}{YanqiDai/VisATB}.
\end{abstract}




\maketitle

\begin{figure}[t!]
    \centering
    \begin{subfigure}[b]{0.93\linewidth}
        \centering
        \includegraphics[width=\linewidth]{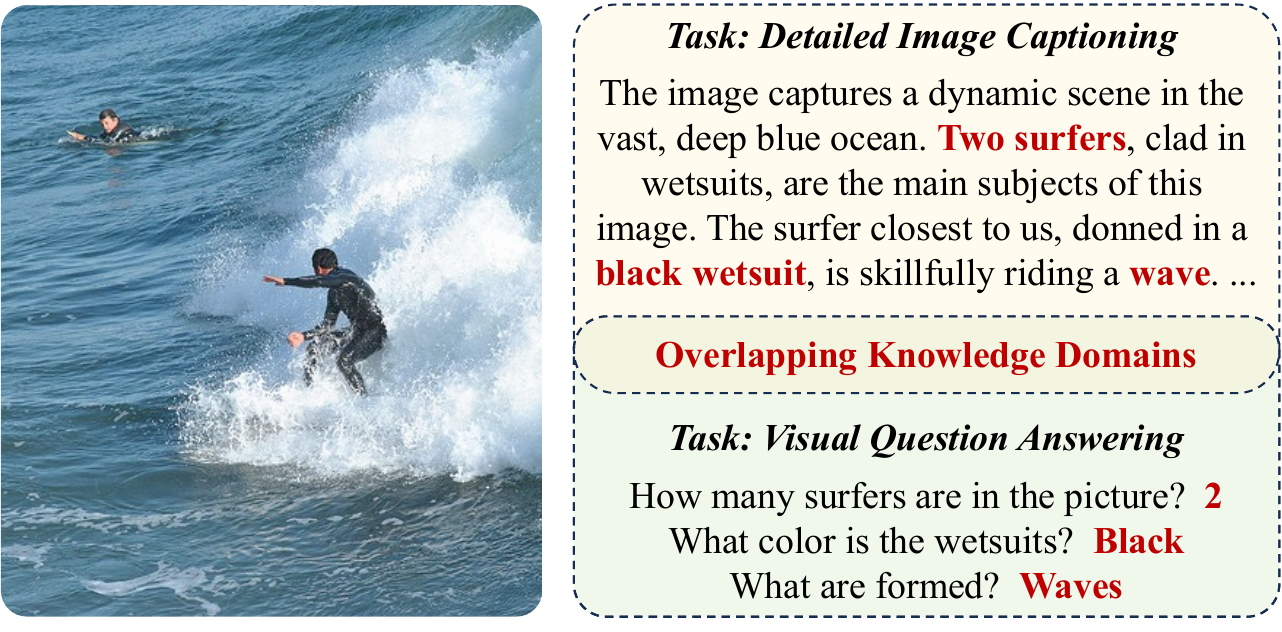}
        \caption{Inter-Task Contribution}
        \label{fig:contribution}
    \end{subfigure}
    \begin{subfigure}[b]{0.99\linewidth}
        \centering
        \includegraphics[width=\linewidth]{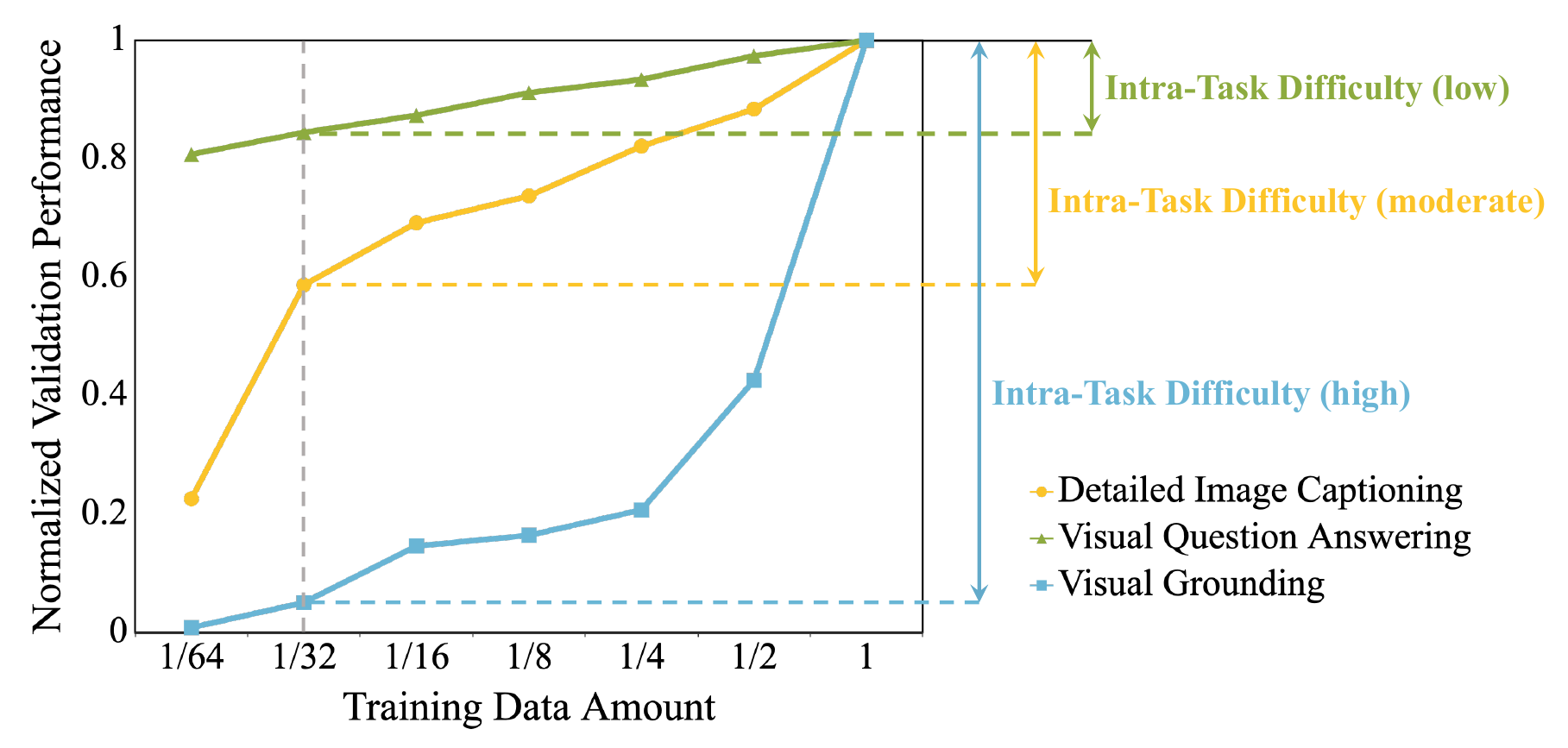}
        \caption{Intra-Task Difficulty}
        \label{fig:difficulty}
    \end{subfigure}
    \vspace{-0.1in}
    \caption{Schematic illustrations of inter-task contributions and intra-task difficulties. (a) The red words reveal that different tasks have overlapping knowledge domains, enabling inter-task contributions. (b) The different performance improvement trajectories w.r.t. training data amount reflect distinct degrees of intra-task difficulties.}
    \label{fig:visatb}
    \vspace{-0.15in}
\end{figure}

\vspace{-0.03in}
\section{Introduction}
\label{sec:intro}
\vspace{-0.01in}

Large multimodal models (LMMs) \citep{yin2024survey, openai2023gpt4, liu2024llavanext} have garnered significant attention for their capability to understand and reason across both visual and textual modalities.
A pivotal advancement in this field is visual instruction tuning \citep{liu2024visual}, which integrates visual encoders with large language models (LLMs) using visual instruction-following data and specialized alignment modules.
This innovative technique extends the robust, general-purpose capabilities of LLMs to the visual modality, substantially enhancing both the efficiency and effectiveness of LMM training.
Many approaches, such as the LLaVA series \citep{liu2024visual, liu2024improved, liu2024llavanext, li2024llavaoneversion}, have achieved remarkable results through visual instruction tuning.

\begin{figure*}[t!]
    \centering
    \includegraphics[width=0.96\linewidth]{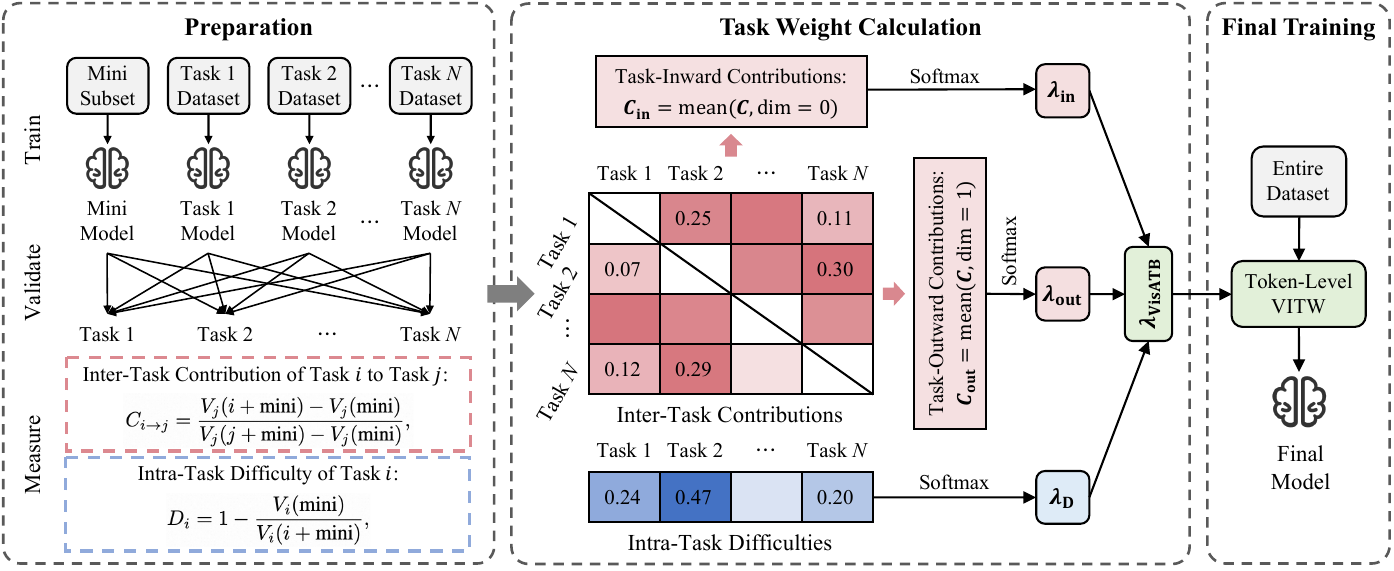}
    \vspace{-0.1in}
    \caption{\textbf{Overview of VisATB.} 
    In the preparation stage, we train models on the mini subset of all tasks and the dataset of each task, and validate their performance across all tasks to measure inter-task contribution and intra-task difficulty.
    In the task weight calculation stage, we compute three types of task weights and integrate them into the task weight $\bm{\lambda_{\textbf{VisATB}}}$.
    In the final training stage, we utilize the entire dataset of all tasks and $\bm{\lambda_{\textbf{VisATB}}}$ to obtain the final model under the VITW paradigm.}
    \label{fig:method}
    \vspace{-0.15in}
\end{figure*}

To endow LMMs with diverse visual abilities, instruction-following data from multiple visual tasks are frequently combined indiscriminately for visual instruction tuning \citep{bai2023qwenvl}.
However, this approach faces a critical challenge: different tasks necessitate task-specific latent knowledge.
For instance, the image captioning task demands comprehensive scene understanding and holistic description generation, whereas the visual grounding task emphasizes fine-grained localization of specific textual phrases.
Consequently, simultaneous training across multiple tasks imposes a substantial learning burden on the model due to \textit{task interference}, potentially resulting in suboptimal performance compared to training on each task individually \citep{gou2023mixture, chen2024llavamole}.
Manual specification of data mixing ratios or task weights heavily relies on extensive ablation studies and expert knowledge, rendering such approaches both resource-intensive and difficult to generalize across different model architectures and tasks.

To mitigate this issue, we perform a rigorous analysis of the task relationship and identify two key concepts: \textbf{inter-task contribution} and \textbf{intra-task difficulty}. First, as illustrated in Figure~\ref{fig:contribution}, we observe that tasks often share overlapping knowledge domains, facilitating knowledge transfer that boosts performance in related tasks.  
The extent of these overlaps varies across tasks, leading to differing degrees of inter-task contributions.
Second, Figure~\ref{fig:difficulty} reflects that different tasks demonstrate distinct performance improvement trajectories as training data increases. 
Specifically, tasks that achieve near-optimal performance with limited training data are regarded as relatively simple, whereas tasks that require extensive training data to reach optimal performance exhibit higher inherent learning difficulty.

Moreover, we introduce a novel \textbf{A}daptive \textbf{T}ask \textbf{B}alancing approach for \textbf{vis}ual instruction tuning (\textbf{VisATB}) based on the above two critical perspectives, integrating three task weighting strategies, each with its unique characteristics, as presented in Figure~\ref{fig:method}.
\textbf{In the preparation stage}, we measure the inter-task contribution of one task to another task by training a model on one task and evaluating its normalized validation performance on the other task.
Additionally, to quantify the intra-task difficulty of a target task, we estimate the normalized validation performance gap between a model trained on a mini subset of the task and one trained on the entire dataset or a sufficiently large subset.
Subsequently, \textbf{in the task weight calculation stage}, we recommend assigning greater weight to tasks that (1) offer substantial contribution to others, (2) receive minimal contribution from others, and (3) present high learning difficulties.
Among these three task weighting strategies, the first and third focus on improving overall performance, while the second aims to mitigate performance imbalance across tasks.
\textit{VisATB integrates them to achieve a more robust and balanced overall performance.}
\textbf{In the final training stage}, we propose a Visual Instruction Task Weighting (VITW) paradigm tailored for visual instruction tuning, where losses are assigned task-specific weights and averaged at the token level.
Building upon this paradigm, the final model is trained using the entire dataset of all tasks and the integrated task weight.

Our contributions can be summarized as follows:
\begin{enumerate}[leftmargin=16pt, topsep=0pt, itemsep=0pt, partopsep=0pt]
    \item We identify two key concepts for visual task balancing: inter-task contribution and intra-task difficulty, and measure them based on validation performance.
    \item We introduce an Adaptive Task Balancing approach for visual instruction tuning (VisATB), which employs three distinct yet complementary task weighting strategies. 
    \item We design a Visual Instruction Task Weighting (VITW) paradigm tailored for visual instruction tuning.
    \item Experiments on three benchmarks indicate that VisATB consistently outperforms existing approaches, achieving a more robust and balanced overall performance.
\end{enumerate}

\vspace{-0.03in}
\section{Method}
\label{sec:method}
\vspace{-0.01in}

In this section, we first introduce the Visual Instruction Task Weighting (VITW) paradigm tailored for visual instruction tuning.
Building upon this paradigm, we analyze two crucial dimensions for visual task balancing: inter-task contribution and intra-task difficulty, and accordingly propose three task weighting strategies. 
Finally, we integrate these strategies to formulate the Adaptive Task Balancing (VisATB) approach.

\vspace{-0.03in}
\subsection{Visual Instruction Task Weighting}
\label{sec:vitw}
\vspace{-0.01in}

Typically, data from various instruction-following tasks are mixed indiscriminately for visual instruction tuning. 
In this process, the training loss is computed as the average cross-entropy loss across all valid tokens, as expressed in the following equation:
\begin{equation}\label{eq:ew}
    L = \frac{\sum^{N}_{i=1}\sum^{S_i}_{j=1}\sum^{T_{ij}}_{k=1}-\log(p(t_{ijk}))}{\sum^{N}_{i=1}\sum^{S_i}_{j=1}T_{ij}},
\end{equation}
where $N$ denotes the task number, $S_i$ indicates the sample number of Task $i$, $T_{ij}$ represents the valid token number in the $j$-th sample for Task $i$, and $t_{ijk}$ signifies the $k$-th valid token in the $j$-th sample for Task $i$.
However, the traditional task weighting paradigm, which computes the total loss as a weighted average of individual task losses, is incompatible in this context.
Differences in sequence length and sample size across tasks result in varying numbers of valid tokens for each task.
Therefore, computing each task loss and averaging across all tasks introduces implicit weight to the losses of valid tokens, leading to biased learning across tasks. 

To address this issue, we design a Visual Instruction Task Weighting (VITW) paradigm tailored for visual instruction tuning.
The training loss of VITW is calculated as:
\begin{equation}
    L_{\text{VITW}} = \frac{\sum^{N}_{i=1}\sum^{S_i}_{j=1}\sum^{T_{ij}}_{k=1}-\lambda_i\log(p(t_{ijk}))}{\sum^{N}_{i=1}\sum^{S_i}_{j=1}\lambda_i T_{ij}},
\end{equation}
where $\lambda_i$ signifies the weight of Task $i$.
The losses of valid tokens are assigned task-specific weight and averaged at the token level, rather than at the task level, to ensure an equitable consideration for each valid token.
It is a robust foundation for VisATB and has the potential to inform future research in visual instruction tuning.

\vspace{-0.03in}
\subsection{Inter-Task Contribution Balancing}
\label{sec:itc}
\vspace{-0.01in}

Although the focal points of different tasks may vary in visual instruction tuning, a central shared objective exists: enhancing the capability of understanding and reasoning about visual information.
As presented in Figure~\ref{fig:contribution}, the detailed image captioning data from ShareGPT4V \citep{chen2023sharegpt4v} and the visual question answering data from VQAv2 \citep{goyal2017making} contain shared information for the same image, such as color, quantity, and category, exemplifying the overlapping knowledge domains among tasks.
Therefore, learning one task can potentially enhance performance on others, through the mechanism we define as inter-task contribution, supported by the results in Section~\ref{sec:exp}.
The extent of inter-task contribution varies according to the degree of overlap in knowledge domains among tasks.

In practice, the inter-task contribution of Task $i$ to Task $j$ can be quantified as the normalized validation performance on Task $j$ of a model trained on Task $i$, as follows:
\begin{equation}
    C_{i\rightarrow j} = \frac{V_{j}(i + \text{mini}) - V_{j}(\text{mini})}{V_{j}(j + \text{mini}) - V_{j}(\text{mini})},
\end{equation}
where $V_{j}(i + \text{mini})$ denotes the performance on Task $j$ of a model trained on the entire dataset or a large enough subset of Task $i$ and mini subsets of all other tasks, and $V_{j}(\text{mini})$ indicates the performance on Task $j$ of a model trained on mini subsets of all tasks.
The large enough subsets are randomly sampled from the entire datasets, while the mini subsets are randomly sampled from these large enough subsets.
To ensure fairness across tasks, each of these two sampling rates remains consistent across all tasks and is independent of the other.
Notably, the mini subsets are added to the training data to ensure the model understands the instruction demands of all tasks.
In the formula, $V_{j}(\text{mini})$ is subtracted from both the numerator and the denominator, which mitigates the influence of incorporating mini subsets into the training set on the validation performance on Task $j$.

Based on the precise quantification of the inter-task contribution, we propose two novel task weighting strategies for inter-task contribution balancing:

\textbf{(1) Task-Outward Contribution Balancing:}
We describe the task-outward contribution, $C_{\text{out}}$, as the average inter-task contribution of a single task to all other tasks.
This concept denotes the extent to which one task contributes to the performance of all other tasks.
Tasks with higher $C_{\text{out}}$ are more beneficial for overall training.
Therefore, we propose assigning greater weight to tasks with higher $C_{\text{out}}$ to improve collective performance across all tasks.
Specifically, the task weight, $\bm{\lambda_{\textbf{out}}}$, for task-outward contribution balancing is computed as:
\begin{equation}
    \bm{\lambda_{\textbf{out}}} = N \times \operatorname{softmax}\left(\frac{\bm{C_{\textbf{out}}}}{T}\right),~\text{where}~C_{\text{out}, i} = \frac{\sum_{j \neq i}C_{i\rightarrow j}}{N-1}.
\label{eq:out}
\end{equation}
Here, $C_{\text{out}, i}$ represents the task-outward contribution of Task $i$, $\bm{C_{\textbf{out}}}$ denotes the task-outward contribution vector of all tasks, and $T$ is the temperature hyperparameter.   

\textbf{(2) Task-Inward Contribution Balancing:} 
Conversely, we characterize the task-inward contribution, $C_{\text{in}}$, as the average inter-task contribution from all other tasks to a single task.
This concept signifies the degree to which the performance of one task benefits from all other tasks.
In comparison to tasks that benefit from more collaborative training, those with lower $C_{\text{in}}$ are more likely to exhibit reduced performance.
Therefore, we propose assigning greater weight to tasks with lower $C_{\text{in}}$ to mitigate performance imbalance.
Specifically, the task weight, $\bm{\lambda_{\textbf{in}}}$, for task-inward contribution balancing is calculated as:
\begin{gather}
    \bm{\lambda_{\textbf{in}}} = N \times \operatorname{softmax}\left(-\frac{\bm{C_{\textbf{in}}}}{T}\right),~\text{where}~C_{\text{in}, i} = \frac{\sum_{j \neq i}C_{j\rightarrow i}}{N-1}.
\label{eq:in}\end{gather}   
Here, $C_{\text{in}, i}$ denotes the task-inward contribution of Task $i$, $\bm{C_{\textbf{in}}}$ signifies the task-inward contribution vector of all tasks, and $T$ is the temperature hyperparameter.

\vspace{-0.03in}
\subsection{Intra-Task Difficulty Balancing}
\label{sec:itd}
\vspace{-0.01in}

Moreover, the inherent learning difficulty, termed intra-task difficulty, also varies substantially across tasks in visual instruction tuning.
As presented in Figure~\ref{fig:difficulty}, different tasks exhibit distinct performance improvement trajectories w.r.t. increasing training data amount.
Tasks that achieve near-optimal performance with limited training data are considered to have lower intra-task difficulties, whereas tasks that require extensive training data to reach optimal performance demonstrate higher intra-task difficulties.
For example, the tasks shown in Figure~\ref{fig:difficulty} are arranged in ascending order of intra-task difficulty as follows: visual question answering, detailed image captioning, and visual grounding.

In practice, the intra-task difficulty of Task $i$ can be measured as the normalized validation performance gap on Task $i$ between a model trained on a mini subset of Task $i$ and one trained on the entire dataset or a large enough subset of the same task.
Additionally, we repurpose the additional models trained for inter-task contribution balancing to reduce time costs.
Consequently, the intra-task difficulty of Task $i$ can be calculated as follows:
\begin{equation}
    D_{i} = 1 - \frac{V_{i}(\text{mini})}{V_{i}(i + \text{mini})},
\end{equation}
where $V_{i}(i + \text{mini})$ indicates the performance on Task $i$ of a model trained on the entire dataset or a large enough subset of Task $i$ and mini subsets of all other tasks, and $V_{i}(\text{mini})$ represents the performance on Task $i$ of a model trained on mini subsets of all tasks.
Importantly, the impact of mini subsets from other tasks on the Task $i$ validation performance is negligible compared to the impact of its own mini subset.
Therefore, model repurposing can significantly reduce additional time costs while maintaining minimal error in the computation of intra-task difficulties.

Due to the varying degrees of intra-task difficulties across tasks, learning all tasks equally may result in underfitting on more challenging tasks, even when simpler ones are overfitted.
Therefore, we propose assigning greater weight to tasks with higher $D$.
Specifically, the task weight, $\bm{\lambda_{\textbf{D}}}$, for intra-task difficulty balancing is computed as follows:
\begin{equation}\label{eq:dif}
    \bm{\lambda_{\textbf{D}}} = N \times \operatorname{softmax}(\frac{\bm{D}}{T}),
\end{equation}
where $\bm{D}$ denotes the intra-task difficulty vector of all tasks, and $T$ is the temperature hyperparameter.

\vspace{-0.03in}
\subsection{VisATB: Adaptive Task Balancing}
\label{sec:visatb}
\vspace{-0.01in}

Finally, these three aforementioned task weighting strategies are integrated to formulate our VisATB approach.
The specific task weight, $\bm{\lambda_{\textbf{VisATB}}}$, for adaptive task balancing is computed as:
\begin{equation}
    \bm{\lambda_{\textbf{VisATB}}} = \alpha_{\text{out}}\bm{\lambda_{\textbf{out}}} + \alpha_{\text{in}}\bm{\lambda_{\textbf{in}}} + \alpha_{\text{D}}\bm{\lambda_{\textbf{D}}},
\end{equation}
where $\alpha_{\text{out}}$, $\alpha_{\text{in}}$, and $\alpha_{\text{D}}$ denote the proportional coefficients, satisfying the constraint $\alpha_{\text{out}} + \alpha_{\text{in}} + \alpha_{\text{D}} = 1$.

In summary, tasks that offer substantial inter-task contributions to others and present high intra-task difficulties are assigned greater weight to achieve superior overall performance, while tasks that receive minimal inter-task contributions from others are assigned greater weight to mitigate performance imbalance.

\begin{table*}[t!]
    \caption{Comparative results on the M$\bm{^3}$IT Benchmark. $\bm{\Delta I\%}$ and $\bm{\Delta E\%}$ are the average per-task improvement and error on fine-tuned tasks compared to the STL baseline.}
    \label{tab:large}
    \vspace{-0.1in}
    \centering
    \scalebox{0.96}{
    \tabcolsep3.2pt
    {\renewcommand{\arraystretch}{1.1}
    \begin{tabular}{l|ccc|cccccc|ccc|ccc|cc|cc}
        \toprule[1.2pt]
        \multirow{3}{*}{Methods} & \multicolumn{3}{c|}{Image Captioning} & \multicolumn{6}{c|}{Classification} & \multicolumn{3}{c|}{VQA} & \multicolumn{3}{c|}{Reasoning} & \multicolumn{2}{c|}{Generation} & \multicolumn{2}{c}{\raisebox{-0.55\normalbaselineskip}[0pt][0pt]{\textbf{Overall}}} \\
        \multirow{3}{*}{ } & COCO & TCap & PCap & GOI & Text & INet & ITM & SVE & Moch & Shap & OCR & GQA & SQA & CLE & NL & VisD & M30k & & \\
        \multirow{3}{*}{ } & \multicolumn{3}{c|}{CIDEr$\uparrow$} & \multicolumn{6}{c|}{EM$\uparrow$} & \multicolumn{3}{c|}{EM$\uparrow$} & \multicolumn{3}{c|}{EM$\uparrow$} & EM$\uparrow$ & GLEU$\uparrow$ & \bm{$\Delta I$}\textbf{\%}\bm{$\uparrow$} & \bm{$\Delta E$}\textbf{\%}\bm{$\downarrow$} \\
        \midrule
        STL & 1.024 & 0.751 & 0.179 & 90.7 & 100 & 98.0 & 98.9 & 81.2 & 84.6 & 63.7 & 56.5 & 48.2 & 73.4 & 57.4 & 72.4 & 42.8 & 0.358 & & \\
        \midrule
        EW & \textbf{1.022} & 0.747 & \underline{0.250} & 91.2 & 100 & \textbf{98.0} & \underline{99.3} & \underline{81.8} & \underline{85.6} & 66.0 & 54.9 & 52.4 & 79.6 & 59.9 & \underline{71.1} & \underline{42.5} & \textbf{0.351} & \underline{3.51} & \underline{0.49} \\
        TLA & 1.001 & 0.739 & 0.219 & \textbf{92.1} & 100 & 97.7 & 99.2 & 81.6 & \underline{85.6} & 65.3 & \textbf{55.5} & 49.5 & 76.5 & \underline{60.4} & 69.0 & 42.2 & 0.342 & 1.43 & 0.97 \\
        RLW & 1.016 & 0.748 & 0.245 & 90.8 & 100 & 97.4 & 99.2 & \underline{81.8} & 79.8 & 64.5 & 54.5 & 50.3 & 76.4 & 55.1 & 69.2 & 41.6 & 0.340 & 1.21 & 1.59  \\
        DWA & \underline{1.021} & \textbf{0.757} & 0.243 & 89.8 & 100 & 97.5 & \underline{99.3} & 81.5 & 79.0 & \underline{67.5} & \textbf{55.5} & \textbf{53.4} & \underline{80.6} & 59.9 & 70.9 & 42.4 & 0.351 & 3.08 & 0.91  \\
        IGBv1 & 1.006 & 0.727 & 0.227 & 89.7 & 100 & 97.1 & 99.1 & 76.1 & 76.4 & 63.9 & 50.6 & 45.4 & 72.2 & 52.8 & 63.0 & 40.8 & 0.343 & -2.55 & 4.16  \\
        \rowcolor{lightblue} VisATB & 1.019 & \underline{0.756} & \textbf{0.258} & \underline{91.7} & 100 & \underline{97.9} & \textbf{99.4} & \textbf{82.2} & \textbf{86.1} & \textbf{68.0} & 55.1 & \underline{52.7} & \textbf{81.1} & \textbf{61.8} & \textbf{71.6} & \textbf{42.7} & \textbf{0.351} & \textbf{4.52} & \textbf{0.39} \\
        \bottomrule[1.2pt]
    \end{tabular}}}
    \vspace{-0.05in}
\end{table*}

\begin{table}[t!]
    \caption{The task weights calculated in VisATB on the M$\bm{^3}$IT Benchmark.} 
    \vspace{-0.1in}
    \label{tab:appendix_weights_m3it}
    \centering
    \scalebox{0.96}{
    \tabcolsep6.4pt
    {\renewcommand{\arraystretch}{1.1}
    \begin{tabular}{l|ccccc}
        \toprule[1.2pt]
        Task Weights & Cap. & Cls. & VQA & Reas. & Gen. \\
        \midrule
        $\bm{\lambda_{\textbf{out}}}$ & 0.9072 & 0.9978 & 1.1010 & 0.9927 & 1.0013 \\
        $\bm{\lambda_{\textbf{in}}}$ & 0.9626 & 0.7375 & 1.2822 & 1.0286 & 0.9891 \\
        $\bm{\lambda_{\textbf{D}}}$ & 1.1831 & 0.9421 & 0.9383 & 0.9967 & 0.9399 \\
        \rowcolor{lightblue} $\bm{\lambda_{\textbf{VisATB}}}$ & 1.0590 & 0.9048 & 1.0649 & 1.0037 & 0.9675 \\
        \bottomrule[1.2pt]
    \end{tabular}}}
    \vspace{-0.05in}
\end{table}

\vspace{-0.03in}
\section{Experiments}
\label{sec:exp}
\vspace{-0.01in}

\subsection{Experimental Setup}
\label{sec:experimental_setup}
\vspace{-0.01in}

\noindent\textbf{Benchmarks.}~~We train and evaluate LMMs using three diverse multimodal benchmarks: a M$^3$IT Benchmark \citep{li2023m3it} comprising 17 training tasks and 1.2 million training samples, an Academic Benchmark \citep{liu2024improved} including 8 training tasks, and a Chat Benchmark \citep{liu2024visual} encompassing 3 training tasks.
Moreover, we assess models on 7 unseen zero-shot tasks in the Academic Benchmark.
A detailed description of the tasks and data preparation for these benchmarks is provided in Appendix~\ref{sec:appendix_data}.

\noindent\textbf{Compared Methods.}~~We compare VisATB against the following baselines and methods:
(1) Single-Task Learning (STL), where models are trained and tested on each single task;
(2) Equal Weighting (EW), the most common approach, which minimizes the loss in Equation~\ref{eq:ew};
(3) Task-Level Loss Aggregation (TLA), which calculates the loss within each task and then averages across all tasks;
(4) Random Loss Weighting (RLW) \citep{lin2021reasonable};
(5) Dynamic Weight Average (DWA) \citep{liu2019end};
and (6) Improvable Gap Balancing (IGBv1) \citep{dai2023improvable}.
Methods (4)-(6) are traditional task weighting methods, as described in Section~\ref{sec:related_work}.
We leverage VITW to adapt them for visual instruction tuning.
Additionally, gradient-based task weighting methods are excluded for comparison due to the substantial computational cost of aggregating gradients from large-scale model parameters.

\noindent\textbf{Evaluation Metrics.}~~We first report the common evaluation metrics for each visual task.
Furthermore, we introduce two overall metrics: 
$\bm{\Delta I}$\textbf{\%}, average per-task improvement, and $\bm{\Delta E}$\textbf{\%}, average per-task error, in test performance on fine-tuned tasks compared to the STL baseline, which are calculated as follows:
\begin{gather}
    \Delta I\% = \frac{1}{N}\sum_{i=1}^{N}I_i,~\Delta E\% = \frac{1}{N}\sum_{i=1}^{N}\max(0, - I_i),\nonumber\\
    \text{where}~I_i = \frac{1}{K_{i}}\sum_{j=1}^{K_{i}} (-1)^{\delta_{ij}}\frac{M_{\text{e},ij} - M_{\text{b},ij}}{M_{\text{b},ij}}.
\end{gather}
Here, $N$ represents the task number, $K_i$ signifies the metric number for Task $i$, and $I_i$ denotes the test performance improvement on Task $i$. 
$M_{\text{e},ij}$ and $M_{\text{b},ij}$ are the values on the $j$-th metric for Task $i$ of the models trained by the evaluated method and the STL baseline.
$\delta_{ij}$ is an indicator function, where $\delta_{ij}=0$ if a higher value is better on the $j$-th metric for Task $i$, and $\delta_{ij}=1$ otherwise.
$\Delta I\%$ reflects the extent of overall performance improvement, while $\Delta E\%$ signifies the degree of performance imbalance.
Moreover, to evaluate the generalizability of methods, we introduce $\bm{\Delta I_{\text{zero}}}$\textbf{\%}, average per-task improvement in test performance on zero-shot tasks compared to EW.
The calculation of $\Delta I_{\text{zero}}\%$ is similar to that of $\Delta I\%$, except the STL baseline is replaced with the EW method.

\noindent\textbf{Implementation Details.}~~In the main experiments, we train the pretrained LLaVA-v1.5-7B model on eight NVIDIA A100 GPUs using the same settings as \citet{liu2024improved}.
This choice is driven by LLaVA's well-established architecture and the accessibility of its pretrained models.
In the M$^3$IT Benchmark, tasks are categorized into 5 groups following \citet{li2023m3it}.
We consider each task group as a whole and perform balancing at the group level.
The temperature $T$ is set as $1.0$ in the M$^3$IT Benchmark, and $0.5$ in the Academic Benchmark and the Chat Benchmark.
The proportional coefficients of three task weighting strategies are consistently set as: $\alpha_{\text{out}} = 0.25$, $\alpha_{\text{in}} = 0.25$, and $\alpha_{\text{D}} = 0.5$.

To calculate inter-task contributions and intra-task difficulties, the entire datasets and $1/32$nd mini subsets of tasks are utilized in the Academic Benchmark, which is guided by the training loss decline pattern, as detailed in Appendix~\ref{sec:appendix_sampling_rate}.
In the Chat Benchmark, there are no specific constraints on the output format.
Consequently, only the entire datasets of tasks are required, without the need for mini subsets, simplifying the form of VisATB, as detailed in Appendix~\ref{sec:appendix_simple_visatb}.
In the M$^3$IT Benchmark, the training data size of each task group is substantial; thus, we use $1/4$th (large enough) subsets and $1/32$nd mini subsets of tasks. 
In practice, guided by experimental analysis in Appendix~\ref{sec:appendix_sampling_rate_large}, we suggest that subsets containing more than 10k samples and trained for over 100 steps are sufficiently large to ensure effective training of VisATB.

\vspace{-0.03in}
\subsection{Evaluation on the M$^3$IT Benchmark}
\label{sec:m3it}
\vspace{-0.01in}

\noindent\textbf{Effectiveness of VisATB.}~~The comparative results of the fine-tuned tasks on the M$^3$IT Benchmark are presented in Table~\ref{tab:large}.
Utilizing the same pretrained model and training data, VisATB achieves the optimal performance in both $\Delta I\%$ and $\Delta E\%$.
Specifically, VisATB attains the best performance on 11 out of 17 tasks and exhibits near-optimal performance on the remaining tasks, which demonstrates the effectiveness of VisATB in both improving overall performance and mitigating performance imbalance across diverse visual tasks.
Notably, in the reasoning group, VisATB substantially outperforms all baselines, such as 81.1 on SQA and 61.8 on CLE, highlighting its potential strength in complex reasoning tasks.

Additionally, the specific task weights calculated in VisATB are presented in Table~\ref{tab:appendix_weights_m3it}.
In the classification task group, although the tasks are assigned relatively small weights (0.9048), VisATB still demonstrates strong performance, attaining the optimal results on 4 out of 6 tasks and ranking second-best on the remaining tasks.
This counterintuitive finding suggests an important insight: for certain relatively simple tasks, directly increasing their weight may not lead to further performance improvements due to potential overfitting or saturation effects.
Instead, our approach of prioritizing tasks that contribute more significantly to these simple ones (as reflected in high task-outward contributions) has the potential to transcend individual task performance limits by facilitating the acquisition of new, transferable knowledge across the task space.

\begin{table*}[t!]
    \caption{Comparative results of the fine-tuned tasks on the Academic Benchmark. $\bm{\Delta I\%}$ and $\bm{\Delta E\%}$ are the average per-task improvement and error on fine-tuned tasks compared to the STL baseline.}
    \vspace{-0.1in}
    \label{tab:results}
    \centering
    \scalebox{0.96}{
    \tabcolsep10pt
    {\renewcommand{\arraystretch}{1.1}
    \begin{tabular}{l|ccccccc|cc}
        \toprule[1.2pt]
        \multirow{2}{*}{Methods} & ShareGPT4V & Ref-caption & VQAv2 & GQA & ChartQA & OCRVQA & Ref-bbox & \multicolumn{2}{c}{\textbf{Overall}} \\
        \multirow{2}{*}{ } & CIDEr$\uparrow$ & CIDEr$\uparrow$ & EM$\uparrow$ & EM$\uparrow$ & EM$\uparrow$ & EM$\uparrow$ & IoU$\uparrow$ & \bm{$\Delta I$}\textbf{\%}\bm{$\uparrow$} & \bm{$\Delta E$}\textbf{\%}\bm{$\downarrow$} \\
        \midrule
        STL & 0.1285 & 0.4658 & 77.73 & 61.23 & 17.76 & 68.22 & 51.58 & & \\
        \midrule
        EW & \underline{0.1411} & 0.5591 & \underline{78.27} & \textbf{62.20} & 19.60 & 67.73 & \underline{61.63} & \underline{8.75} & \textbf{0.10}\\
        TLA & 0.1144 & \textbf{0.5770} & 77.72 & 60.42 & \textbf{22.36} & \underline{67.80} & 56.58 & 6.65 & 1.85\\
        RLW & 0.1388 & 0.5571 & 77.28 & 60.61 & 18.20 & 66.73 & 55.86 & 4.95 & 0.54 \\
        DWA & 0.1225 & 0.5470 & \textbf{78.28} & \underline{61.82} & 19.88 & \textbf{67.87} & 61.12 & 6.34 & 0.74 \\
        IGBv1 & 0.1349 & 0.4824 & 77.00 & 60.92 & 17.20 & 65.96 & 55.47 & 0.17 & 1.13 \\
        \rowcolor{lightblue} VisATB & \textbf{0.1437} & \underline{0.5724} & 77.99 & 61.81 & \underline{20.16} & 67.48 & \textbf{67.38} & \textbf{11.29} & \underline{0.15}\\
        \bottomrule[1.2pt]
    \end{tabular}}}
    \vspace{-0.02in}
\end{table*}

\begin{table*}[t!]
    \caption{Comparative results of the zero-shot tasks on the Academic Benchmark. $\bm{\Delta I_{\text{zero}}\%}$ is the average per-task improvement on zero-shot tasks compared to the EW method.}
    \vspace{-0.1in}
    \label{tab:zero_shot_results}
    \centering
    \scalebox{0.96}{
    \tabcolsep10.6pt
    {\renewcommand{\arraystretch}{1.1}
    \begin{tabular}{l|ccccccc|c}
        \toprule[1.2pt]
        Methods & TextVQA$\uparrow$ & POPE$\uparrow$ & MME$\uparrow$ & SQA$\uparrow$ & MMBench$\uparrow$ & $\text{SEED}^\text{I}$$\uparrow$ & MM-Vet$\uparrow$ & \textbf{Overall (\bm{$\Delta I_{\textbf{zero}}$}\textbf{\%}\bm{$\uparrow$})} \\
        \midrule
        EW & \underline{53.96} & 86.83 & \textbf{1524.41} & \underline{60.74} & 48.54 & 54.13 & 29.00 & 0.00 \\
        TLA & 50.74 & \textbf{86.94} & 1426.51 & 55.86 & \textbf{52.92} & 46.67 & 28.70 & -3.73 \\ 
        RLW & 52.74 & 86.29 & \underline{1504.06} & 59.99 & \underline{52.58} & \underline{56.54} & 28.70 & \underline{0.90} \\
        DWA & \textbf{54.05} & 86.76 & 1484.26 & 60.53 & 49.31 & 54.75 & 28.90 & -0.07 \\
        IGBv1 & 52.09 & \underline{86.86} & 1497.02 & 57.79 & 39.95 & 47.26 & \textbf{29.30} & -5.63 \\
        \rowcolor{lightblue} VisATB & 53.87 & 86.73 & 1501.86 & \textbf{61.07} & 51.46 & \textbf{58.05} & \textbf{29.30} & \textbf{1.87}\\
        \bottomrule[1.2pt]
    \end{tabular}}}
    \vspace{-0.02in}
\end{table*}

\noindent\textbf{Validity of VITW.}~~We compare TLA with EW to systematically evaluate the validity of our VITW paradigm in visual instruction tuning.
The results provide compelling evidence for the necessity of token-level loss aggregation.
Specifically, TLA performs substantially worse than EW in both overall metrics: it achieves only 1.43\% in $\Delta I\%$ compared to 3.51\% of EW, and exhibits a more pronounced performance imbalance with a $\Delta E\%$ of 0.97\% versus 0.49\% of EW.
As discussed in Section~\ref{sec:vitw}, this performance degradation stems from the implicit weight bias introduced by TLA, which is inversely proportional to the number of valid tokens in each task.
For example, the image captioning task group, which naturally involves longer textual descriptions and thus a larger number of valid tokens per sample, receives lower implicit weight in the optimization process, thereby leading to the poorest performance.
These findings underscore the validity of our VITW paradigm, which ensures equitable treatment of all valid tokens by computing weighted losses at the token level before aggregation.

\noindent\textbf{Limitation of Traditional Task Weighting Methods.}~~Directly applying traditional task weighting methods, including RLW, DWA, and IGBv1, in visual instruction tuning yields markedly inferior performance.
These results underscore a fundamental limitation of traditional methods, which balance tasks solely based on training losses, when applied to visual instruction tuning: training losses fail to serve as reliable indicators of actual task learning progress and generalization capacity in LMMs.
In contrast, the validation performance-based measurement in VisATB provides a more accurate reflection of the model's actual capabilities and learning trajectories across tasks, leading to superior task weight determination and significant performance improvements.

\noindent\textbf{Quantitative Analysis of the Time Cost of VisATB.}~~An important practical consideration for any task balancing method is its computational overhead.
We provide a comprehensive analysis of the additional training time required by VisATB.
The extra training time of VisATB is approximately $(R_{\text{large}}+N\times R_{\text{mini}})$ times the training duration of the final model, where $N$ denotes the task number, $R_{\text{large}}$ and $R_{\text{mini}}$ represent the sampling rates of large enough subsets and mini subsets.
In the M$^3$IT Benchmark, with $N=5$, $R_{\text{large}}=1/4$, and $R_{\text{mini}}=1/32$, the additional training time is about $0.25 + 5 \times 0.03 = 0.40$ times that of the final model.

However, this overhead exhibits favorable scaling properties:
\begin{enumerate}[leftmargin=16pt, topsep=2pt, itemsep=0pt, partopsep=0pt]
\item\textit{Sublinear scaling with data size}: As the total dataset size increases, both $R_{\text{large}}$ and $R_{\text{mini}}$ can be proportionally reduced while maintaining sufficient samples for reliable measurement (as discussed in Section~\ref{sec:appendix_sampling_rate_large}, subsets with >10k samples trained for >100 steps are generally sufficient). This ensures that the absolute additional cost does not increase significantly with dataset scale.
\item\textit{Task grouping strategy}: In scenarios with a large number of tasks, e.g., the M$^3$IT Benchmark with 17 tasks, VisATB can leverage task clustering to perform balancing at the group level. By clustering semantically or structurally similar tasks (e.g., all classification tasks into one group), we can reduce $N$ from 17 to 5, substantially decreasing the $(N \times R_{\text{mini}})$ component of the overhead.
It can be performed using expert knowledge about task similarities or automated clustering approaches based on task characteristics.
\end{enumerate}
Consequently, VisATB can effectively function across diverse scales and scenarios without substantially increasing complexity or incurring prohibitive time costs, making it a practical solution for real-world visual instruction tuning applications.

\vspace{-0.03in}
\subsection{Evaluation on the Academic Benchmark}
\label{sec:academic}
\vspace{-0.01in}

The comparative results of the fine-tuned tasks on the Academic Benchmark are summarized in Table~\ref{tab:results}.
Overall, VisATB achieves the highest $\Delta I\%$ while maintaining a near-minimal $\Delta E\%$, indicating both enhanced overall performance and mitigated task imbalance.
Compared with EW, VisATB delivers substantial gains on Ref-bbox, Ref-caption, ChartQA, and ShareGPT4V, while preserving competitive results on the remaining tasks. 
These results further demonstrate the effectiveness of VisATB.
Additionally, TLA and all traditional task weighting methods perform worse than EW and VisATB in both $\Delta I\%$ and $\Delta E\%$, further underscoring the validity of VITW and the limitation of traditional methods.

\noindent\textbf{Generalization on Zero-Shot Tasks.}~~The comparative results of the zero-shot tasks on the Academic Benchmark are presented in Table~\ref{tab:zero_shot_results}, including single-word QA (TextVQA, POPE, MME), multiple-choice QA (SQA, MMBench, $\text{SEED}^\text{I}$), and open-ended QA (MM-Vet).
Overall, VisATB achieves the optimal $\Delta I_{\text{zero}}\%$.
Compared to EW, VisATB exhibits significantly superior performance on multi-choice QA tasks while maintaining comparable results on other tasks.
Importantly, the training corpus contains no visual multiple-choice QA-type data at all.
Nevertheless, VisATB substantially enhances the model's zero-shot generalization to such tasks, demonstrating its ability to induce transferable capabilities beyond the supervised domains.
Additionally, TLA demonstrates markedly lower values than EW in $\Delta I_{\text{zero}}\%$, further supporting the validity of VITW.

\begin{figure*}[t!]
    \centering
    \begin{subfigure}[b]{0.48\linewidth}
        \centering
        \includegraphics[width=\linewidth]{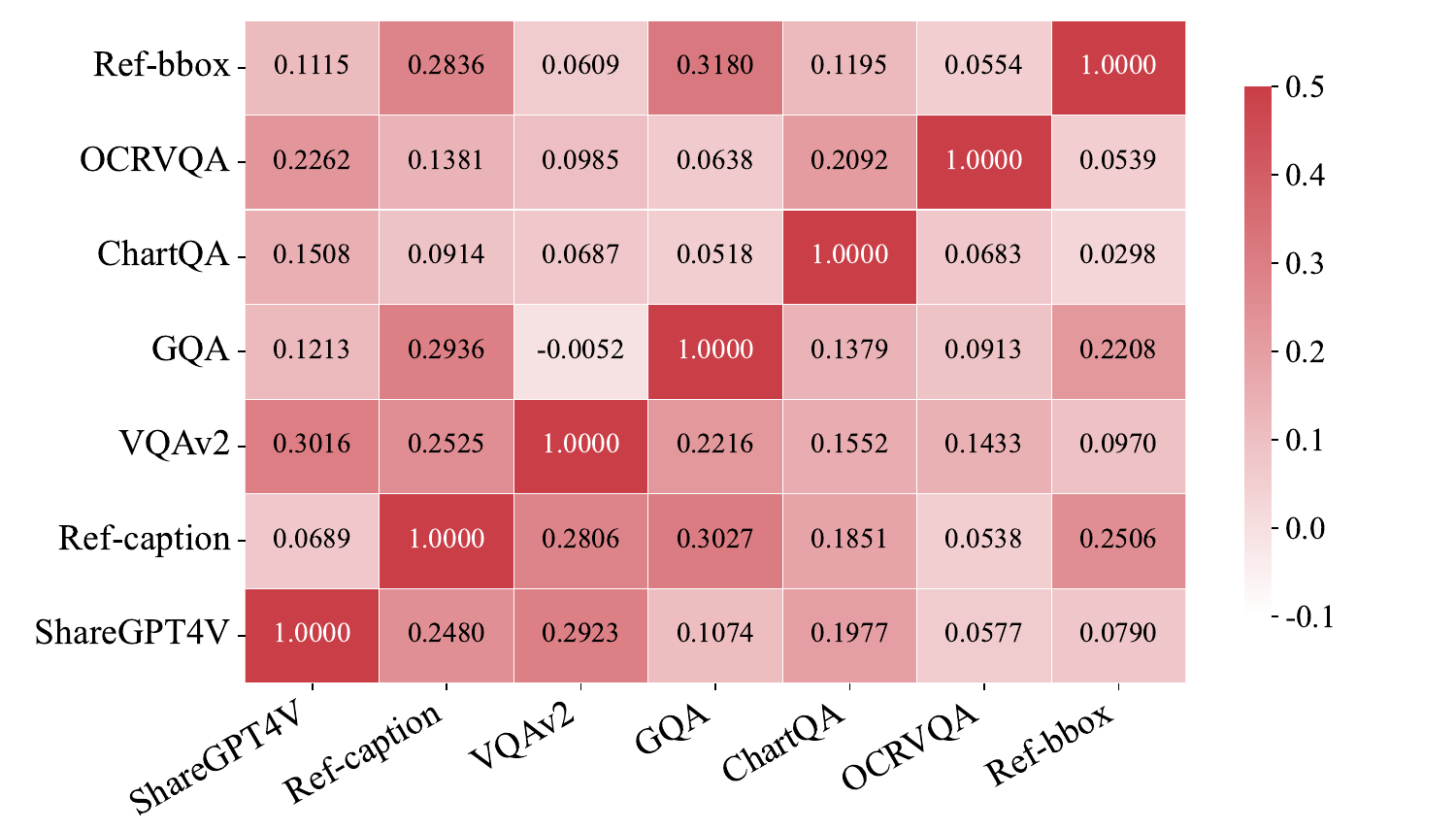}
        \vspace{-0.2in}
        \caption{Heatmap of inter-task contributions.}
        \label{fig:visual_contribution}
    \end{subfigure}
    \begin{subfigure}[b]{0.48\linewidth}
        \centering
        \includegraphics[width=\linewidth]{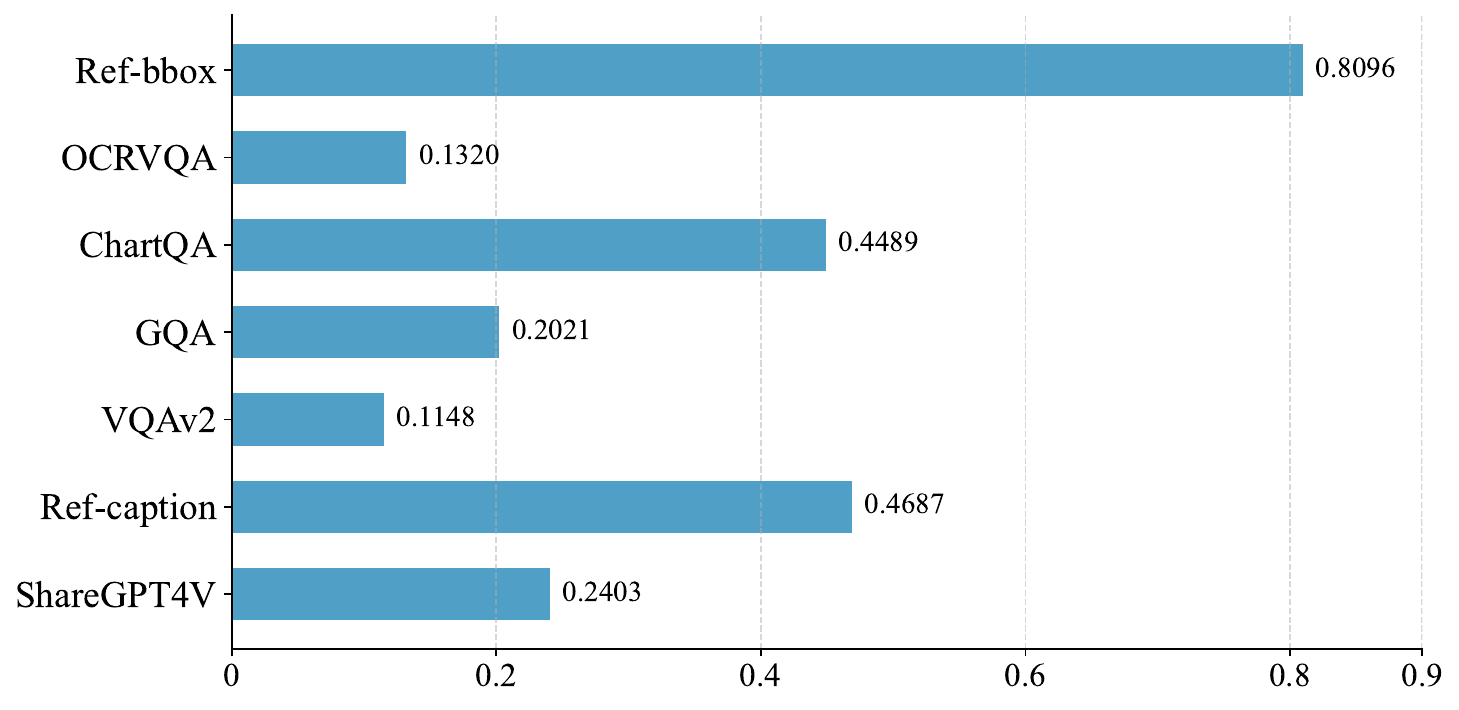}
        \vspace{-0in}
        \caption{Histogram of intra-task difficulties.}
        \label{fig:visual_difficulty}
    \end{subfigure}
    \vspace{-0.1in}
    \caption{Visualizations of inter-task contributions and intra-task difficulties calculated in VisATB on the Academic Benchmark.}
    \label{fig:visual}
    \vspace{-0.05in}
\end{figure*}

\begin{table*}[t!]
    \caption{The task weights calculated in VisATB on the Academic Benchmark.} 
    \vspace{-0.1in}
    \label{tab:appendix_weights}
    \centering
    \scalebox{0.96}{
    \tabcolsep13.8pt
    {\renewcommand{\arraystretch}{1.1}
    \begin{tabular}{l|ccccccc}
        \toprule[1.2pt]
        Task Weights & ShareGPT4V & Ref-caption & VQAv2 & GQA & ChartQA & OCRVQA & Ref-bbox \\
        \midrule
        $\bm{\lambda_{\textbf{out}}}$ & 1.0223 & 1.0782 & 1.0888 & 0.9815 & 0.8592 & 0.9588 & 1.0112 \\
        $\bm{\lambda_{\textbf{in}}}$ & 0.9727 & 0.8722 & 1.0345 & 0.9456 & 0.9649 & 1.1532 & 1.0569 \\
        $\bm{\lambda_{\textbf{D}}}$ & 0.7195 & 1.1361 & 0.5598 & 0.6666 & 1.0920 & 0.5794 & 2.2466 \\
        \rowcolor{lightblue} $\bm{\lambda_{\textbf{VisATB}}}$ & 0.8585 & 1.0557 & 0.8107 & 0.8151 & 1.0020 & 0.8177 & 1.6403 \\
        \bottomrule[1.2pt]
    \end{tabular}}}
    \vspace{-0.03in}
\end{table*}

\begin{table}[t!]
    \caption{Comparative results on the Academic Benchmark using various pretrained models.}
    \vspace{-0.1in}
    \label{tab:model_results}
    \centering
    \scalebox{0.96}{
    \tabcolsep7.2pt
    {\renewcommand{\arraystretch}{1.1}
    \begin{tabular}{l|l|ccc}
        \toprule[1.2pt]
        Models & Methods & \bm{$\Delta I$}\textbf{\%}\bm{$\uparrow$} & \bm{$\Delta E$}\textbf{\%}\bm{$\downarrow$} & \bm{$\Delta I_{\textbf{zero}}$}\textbf{\%}\bm{$\uparrow$} \\
        \midrule
        \multirow{2}{*}{LLaVA-v1.5-13B} & EW & 4.55 & 0.66 & 0.00 \\
        \multirow{2}{*}{} & \cellcolor{lightblue}VisATB & \cellcolor{lightblue}\textbf{6.89} & \cellcolor{lightblue}\textbf{0.29} & \cellcolor{lightblue}\textbf{0.38} \\
        \midrule
        \multirow{2}{*}{Qwen2-VL-2B} & EW & -2.68 & 4.08 & 0.00 \\
        \multirow{2}{*}{} & \cellcolor{lightblue}VisATB & \cellcolor{lightblue}\textbf{-1.32} & \cellcolor{lightblue}\textbf{3.70} & \cellcolor{lightblue}\textbf{1.00} \\
        \bottomrule[1.2pt]
    \end{tabular}}}
    \vspace{-0.05in}
\end{table}

\noindent\textbf{Visual Analysis of VisATB.}~~The inter-task contributions and intra-task difficulties of VisATB are visualized in Figure~\ref{fig:visual}, and the resulting task weights are detailed in Table~\ref{tab:appendix_weights}.
As shown in Figure~\ref{fig:visual_contribution}, the heatmap of inter-task contributions demonstrates substantial heterogeneity across task pairs: certain tasks (e.g., Ref-caption and VQAv2) provide strong positive contributions to others, while some (e.g., OCRVQA) receive minimal benefit from others.
Meanwhile, Figure~\ref{fig:visual_difficulty} illustrates that intra-task difficulties vary significantly: tasks like grounding and captioning are notably harder to learn compared to simpler tasks such as basic visual QA.
This highlights the necessity of considering both inter-task contributions and intra-task difficulties for effective task balancing, and the effectiveness of VisATB to capture these differences across tasks.

\noindent\textbf{Model Independence of VisATB.}~~To verify the general effectiveness of our approach independent of specific models, we further fine-tune the pretrained LLaVA-v1.5-13B \citep{liu2024improved} and Qwen2-VL-2B \citep{wang2024qwen2} models on the Academic Benchmark.
The overall results are presented in Table~\ref{tab:model_results}, with detailed results and settings in Appendix~\ref{sec:appendix_pretrain}.
Across both scales and architectures, VisATB consistently improves $\Delta I\%$ and $\Delta I_{\text{zero}}\%$ while reducing $\Delta E\%$ compared to EW, demonstrating its robustness to backbone variation.

\vspace{-0.03in}
\subsection{Evaluation on the Chat Benchmark}
\vspace{-0.01in}

The comparative results of the fine-tuned tasks on the Chat Benchmark are presented in Table~\ref{tab:chat}, which includes three loosely structured tasks: general conversation (Conv.), detailed description (Detail.), and complex reasoning (Complex.).
This setting reduces overt format conflicts across tasks and serves to assess whether VisATB still offers benefits when structural incompatibilities are minimal.
Overall, VisATB outperforms all methods except TLA in $\Delta I\%$, while maintaining the lowest value in $\Delta E\%$.
Compared to EW, VisATB yields substantial gains on Conv.\ (+7.4 absolute, 52.2 vs. 44.8) and performs slightly better on Detail.\ (60.7 vs. 59.6), while sustaining the state-of-the-art Complex.\ performance (83.4).
These results indicate that in a chat-style setting, VisATB can still enhance underfitting tasks without compromising the performance of well-performing tasks.

\noindent\textbf{Analysis of TLA and Traditional Methods.}~~TLA achieves the best $\Delta I\%$ but at the cost of an increase in $\Delta E\%$ (0.70 vs. 0.00 of EW).
Because the implicit weight introduced by TLA is unregulated, it is unstable across various scenarios and prone to performance imbalance.
Additionally, traditional task weighting methods exhibit inferior performance in $\Delta I\%$, except for DWA, which outperforms EW but still falls short of VisATB.
These findings further indicate the necessity of our VITW paradigm and the insufficiency of traditional methods for visual instruction tuning.

\begin{table}[t!]
    \caption{Comparative results on the Chat Benchmark. $\bm{\Delta I\%}$ and $\bm{\Delta E\%}$ are the average per-task improvement and error on fine-tuned tasks compared to the STL baseline.}
    \vspace{-0.1in}
    \label{tab:chat}
    \centering
    \scalebox{0.96}{
    \tabcolsep6pt
    {\renewcommand{\arraystretch}{1.1}
    \begin{tabular}{l|ccc|cc}
        \toprule[1.2pt]
        \multirow{2}{*}{Methods} & \multirow{2}{*}{Conv.$\uparrow$} & \multirow{2}{*}{Detail.$\uparrow$} & \multirow{2}{*}{Complex.$\uparrow$} & \multicolumn{2}{c}{\textbf{Overall}} \\
        \multirow{2}{*}{} & \multirow{2}{*}{} & \multirow{2}{*}{} & \multirow{2}{*}{} & \bm{$\Delta I$}\textbf{\%}\bm{$\uparrow$} & \bm{$\Delta E$}\textbf{\%}\bm{$\downarrow$} \\
        \midrule
        STL & 43.6 & 49.3 & 80.9 & & \\
        \midrule
        EW & 44.8 & 59.6 & \textbf{83.4} & 8.91 & \textbf{0.00} \\
        TLA & \textbf{57.2} & 61.9 & 79.2 & \textbf{18.22} & 0.70 \\
        RLW & 46.0 & 56.7 & 82.9 & 7.66 & \textbf{0.00} \\
        DWA & 46.1 & \textbf{65.9} & 82.8 & 13.92 & \textbf{0.00} \\
        IGBv1 & 39.9 & \underline{64.2} & 81.7 & 7.58 & 2.83 \\
        \rowcolor{lightblue} VisATB & \underline{52.2} & 60.7 & \textbf{83.4} & \underline{15.31} & \textbf{0.00} \\
        \bottomrule[1.2pt]
    \end{tabular}}}
    \vspace{-0.05in}
\end{table}

\vspace{-0.03in}
\subsection{Ablation Studies}
\vspace{-0.01in}

As presented in Table~\ref{tab:ablation} (with full results provided in Appendix~\ref{sec:appendix_full_ablation}), we ablate our VisATB on the Academic Benchmark from three perspectives: task weighting strategies, temperatures, and calculation approaches for the intra-task difficulty.
The compared methods include: EW; VisATB ($\bm{\alpha}\!=\![\alpha_{\text{out}}, \alpha_{\text{in}}, \alpha_{\text{D}}]$), where the proportional coefficients for the three task weighting strategies are set to different values;
VisATB ($T\!=\!2.0/1.0/0.5$), where the temperature $T$ is set as $2.0$, $1.0$ or $0.5$;
and VisATB (precise/real Diff), where the precise or real calculation approach for intra-task difficulty is used.
The precise calculation approach trains the additional models to precisely quantify the intra-task difficulty, as detailed in Appendix~\ref{sec:appendix_precise},
while the real calculation approach repurposes the models trained for inter-task contribution balancing to reduce the time cost.

\begin{table}[t!]
    \caption{Ablation results on the Academic Benchmark. $\bm{\alpha\!=\![\alpha_{\text{out}}, \alpha_{\text{in}}, \alpha_{\text{D}}]}$ is the vector of proportional coefficients for the three task weighting strategies; $\bm{T}$ is the temperature hyperparameter; and `precise/real Diff' denotes the use of the precise or real calculation approach for intra-task difficulty.}
    \vspace{-0.1in}
    \label{tab:ablation}
    \centering
    \scalebox{0.96}{
    \tabcolsep6.4pt
    {\renewcommand{\arraystretch}{1.1}
    \begin{tabular}{l|ccc}
        \toprule[1.2pt]
        Methods & \bm{$\Delta I$}\textbf{\%}\bm{$\uparrow$} & \bm{$\Delta E$}\textbf{\%}\bm{$\downarrow$} & \bm{$\Delta I_{\textbf{zero}}$}\textbf{\%}\bm{$\uparrow$} \\
        \midrule
        EW & 8.75 & 0.10 & 0.00\\
        \midrule
        VisATB ($\bm{\alpha}\!=\![1,0,0]$) & \underline{9.67} & \underline{0.10} & \textbf{1.36} \\
        VisATB ($\bm{\alpha}\!=\![0,1,0]$) & 8.45 & \textbf{0.05} & -3.96\\
        VisATB ($\bm{\alpha}\!=\![0,0,1]$) & \textbf{12.21} & 0.30 & \underline{0.63} \\
        \midrule
        VisATB ($\bm{\alpha}\!=\![0.50,0.50,0]$) & 8.58 & \textbf{0.06} & \underline{2.32} \\
        VisATB ($\bm{\alpha}\!=\![0.33,0.33,0.33]$) & \underline{8.77} & 0.32 & \textbf{3.04} \\
        \rowcolor{lightblue}VisATB ($\bm{\alpha}\!=\![0.25,0.25,0.50]$) & \textbf{11.29} & \underline{0.15} & 1.87 \\
        \midrule
        VisATB ($T\!=\!2.0$) & 9.92 & \underline{0.12} & 0.50\\
        VisATB ($T\!=\!1.0$) & \underline{10.24} & \textbf{0.11} & \underline{0.52}\\
        \rowcolor{lightblue}VisATB ($T\!=\!0.5$) & \textbf{11.29} & 0.15 & \textbf{1.87}\\
        \midrule
        VisATB (precise Diff) & \underline{10.75} & \underline{0.16} & \textbf{2.61}\\
        \rowcolor{lightblue} VisATB (real Diff) & \textbf{11.29} & \textbf{0.15} & \underline{1.87}\\
        \bottomrule[1.2pt]
    \end{tabular}}}
    \vspace{-0.05in}
\end{table}

\noindent\textbf{Task Weighting Strategies.}~~We first isolate each strategy component of VisATB.
As mentioned in Section~\ref{sec:visatb}, $\bm{\lambda_{\textbf{out}}}$ and $\bm{\lambda_{\textbf{D}}}$ focus on improving overall performance, while $\bm{\lambda_{\textbf{in}}}$ aims to mitigate performance imbalance.
Compared to EW, activating only the $\bm{\lambda_\text{out}}$ weight (VisATB ($\bm{\alpha}\!=\![1,0,0]$)) raises $\Delta I\%$ from $8.75$ to $9.67$ and improves $\Delta I_{\text{zero}}\%$ to $1.36$, while keeping $\Delta E\%$ unchanged at $0.10$.
Utilizing only the $\bm{\lambda_\text{in}}$ weight (VisATB ($\bm{\alpha}\!=\![0,1,0]$)) effectively reduces $\Delta E\%$ to $0.05$, albeit with a slight drop in $\Delta I\%$ and $\Delta I_{\text{zero}}\%$.
Relying solely on the $\bm{\lambda_\text{D}}$ weight (VisATB ($\bm{\alpha}\!=\![0,0,1]$)) produces the largest boost in $\Delta I\%=12.21$ and a moderate improvement in $\Delta I_{\text{zero}}\%$ but at the cost of a slight increase in $\Delta E\%$.
These underscore the effectiveness of all three task weighting strategies, each focusing on distinct yet complementary aspects.

When combining these task weighting strategies, the proportional coefficients can be adjusted to achieve more favorable Pareto trade-offs.
Specifically, VisATB ($\bm{\alpha}\!=\![0.50,0.50,0]$) integrates two inter-task contribution balancing strategies, resulting in a balance between $\Delta I\%$ and $\Delta E\%$, while also enhancing $\Delta I_{\text{zero}}\%$.
Moreover, VisATB ($\bm{\alpha}\!=\![0.33,0.33,0.33]$) combines all three strategies, leading to improvements in $\Delta I\%$ and $\Delta I_{\text{zero}}\%$.
To further enhance overall performance, VisATB ($\bm{\alpha}\!=\![0.25,0.25,0.50]$) slightly increases the value of $\alpha_{\text{D}}$, achieving significantly higher $\Delta I\%$ and $\Delta I_{\text{zero}}\%$ than EW, while also maintaining nearly the lowest $\Delta E\%$.
Notably, we observe that combining strategies generally leads to superior performance in $\Delta I_{\text{zero}}\%$ than employing any single strategy.
This finding demonstrates the effectiveness of comprehensively considering multiple aspects of task balancing for generalization to unseen zero-shot tasks.

\noindent\textbf{Temperatures.}~~VisATB increasingly outperforms EW in both $\Delta I\%$ and $\Delta I_{\text{zero}}\%$ as $T$ decreases, while exhibiting a slightly higher $\Delta E\%$ at lower values of $T$.
As the temperature in the softmax functions of Equations~\ref{eq:out}, \ref{eq:in}, and \ref{eq:dif} decreases, the weight distribution becomes progressively sharper.
If the sharpness in task weight is excessively high, tasks with too small weights may inevitably underperform, leading to a slight performance imbalance.
In practice, we recommend using the lowest temperature $T$ that ensures all task weights remain within the range of $0.5$ to $2.0$ to avoid over-balancing.

\noindent\textbf{Calculation Approaches for Intra-Task Difficulty.}~~As discussed in Section~\ref{sec:itd}, the objective of the real approach is to reduce the time cost with minimal error.
VisATB (precise Diff) and VisATB (real Diff) exhibit comparable performance levels, with VisATB (real Diff) even showing a slight advantage in $\Delta I\%$ and $\Delta E\%$. 
Meanwhile, the real calculation approach enables a reduction of approximately $(R_{\text{large}}+R_{\text{mini}})$ times the training duration of the final model, where $R_{\text{large}}$ and $R_{\text{mini}}$ represent the sampling rates of large enough subsets and mini subsets, respectively.
This observation underscores the efficacy of our real calculation approach.

\vspace{-0.03in}
\section{Related Work}
\label{sec:related_work}
\vspace{-0.01in}

\noindent\textbf{Visual Instruction Tuning.}~~Instruction tuning is first proposed in NLP, enabling LLMs to follow textual instructions and accomplish various tasks \citep{wei2021finetuned}.
Moreover, to extend the powerful abilities of LLMs into the multimodal domain, \citet{liu2024visual} introduce visual instruction tuning.
This innovative technique integrates LLMs with visual encoders using visual instruction-following data and alignment modules. 
Subsequently, a series of improved approaches demonstrate robust performance on visual tasks, respectively focusing on model structures \citep{zhu2023minigpt4, dai2023instructblip, bai2023qwenvl, gou2023mixture, lin2024moellava, shen2024mome, chen2024llavamole}, training settings \citep{liu2024improved, ye2023mplugowl}, and training data \citep{zhao2023svit, zhang2023llavar, li2024llava, chen2023sharegpt4v, wang2024vigc}.
Particularly, to mitigate visual task conflicts, \citet{dai2023instructblip} adaptively adjust sampling probabilities based on task data sizes, and several recent studies design the mixture of LoRA experts structure \citep{gou2023mixture, chen2024llavamole, shen2024mome}.
In this paper, we propose tackling this challenge from an alternative perspective by adaptive task weighting.

\noindent\textbf{Task Weighting.}~~Adaptive task weighting is commonly employed in CV, which assigns task weights based on losses or gradients to balance the joint training process of tasks \citep{sener2018multi, kendall2018multi, liu2021towards, liu2021conflict, navon2022multi, achituve2024bayesian, ban2024fair}.
For example, \citet{lin2021reasonable} assign task weight randomly;
\citet{liu2019end} prefer tasks with lower loss decline rates; 
and \citet{dai2023improvable} favor tasks with higher improvable gaps.

\vspace{-0.03in}
\section{Conclusion}
\vspace{-0.01in}

In this paper, we introduce an Adaptive Task Balancing approach for visual instruction tuning (VisATB).
Specifically, we design a token-level Visual Instruction Task Weighting (VITW) paradigm. 
Building upon this paradigm, we analyze two crucial dimensions for visual task balancing: inter-task contribution and intra-task difficulty.
Accordingly, we propose three distinct yet complementary task weighting strategies.
Extensive experiments demonstrate that VisATB outperforms existing methods, achieving a more robust and balanced overall performance.

\begin{acks}
This work was supported in part by National Natural Science Foundation of China (62376274, 62437002).
\end{acks}

\bibliographystyle{ACM-Reference-Format}
\balance
\bibliography{sample-base}

\appendix

\section{Task Information and Data Preparation}
\label{sec:appendix_data}

We train and evaluate LMMs on the following benchmarks:

\noindent\textbf{M$^3$IT Benchmark.}~~The tasks in the M$^3$IT Benchmark are carefully selected from the M$^3$IT dataset \citep{li2023m3it}, a large-scale multimodal instruction tuning dataset.
All curated tasks have training and validation sets.
If no test set is provided, the validation set is randomly divided into two equal parts, one for validation and the other for testing.
Following the task clustering of \citet{li2023m3it}, the tasks are categorized into five distinct groups.
The task grouping and task list in the group are shown as follows:
\begin{enumerate}[leftmargin=16pt, topsep=0pt, itemsep=1pt, partopsep=0pt]
    \item Image Captioning: MS-COCO (COCO) \citep{lin2014microsoft}, TextCaps (TCap) \citep{sidorov2020textcaps}, and Image-Paragraph-Captioning (PCap) \citep{krause2017hierarchical}.
    \item Classification: COCO-GOI (GOI) \citep{lin2014microsoft}, COCO-Text (Text) \citep{veit2016coco}, ImageNet (INet) \citep{russakovsky2015imagenet}, COCO-ITM (ITM) \citep{lin2014microsoft}, e-SNLI-VE (SVE) \citep{kayser2021vil}, and Mocheg (Moch) \citep{yao2023end}.
    \item Visual Question Answering (VQA): Shapes VQA (Shap) \citep{andreas2016neural}, OCRVQA (OCR) \citep{mishra2019ocr}, and GQA \citep{hudson2019gqa}
    \item Reasoning: ScienceQA (SQA) \citep{lu2022learn}, CLEVR (CLE) \citep{johnson2017clevr}, and NLVR (NL) \citep{suhr2017corpus}.
    \item Generation: Visual Dialog (VisD) \citep{das2017visual}, and Multi30k (M30k) \citep{elliott2016multi30k}.
\end{enumerate}

\noindent\textbf{Chat Benchmark.}~~The tasks in the Chat Benchmark are introduced by \citet{liu2024visual}, including general conversation (Conv.), detailed description (Detail.), and complex reasoning (Complex.).
We utilize LLaVA-Bench-COCO for validation and $\text{LLaVA}^\text{W}$ for testing.

\noindent\textbf{Academic Benchmark.}~~The tasks in the Academic Benchmark encompass ShareGPT \citep{sharegpt}, ShareGPT4V \citep{chen2023sharegpt4v}, Ref-caption, Ref-bbox \citep{kazemzadeh2014referitgame, mao2016generation}, VQAv2 \citep{goyal2017making}, GQA \citep{hudson2019gqa}, ChartQA \citep{masry2022chartqa}, and OCRVQA \citep{mishra2019ocr}.
Among these, the Ref-caption task involves generating captions for image regions defined by bounding boxes, while the Ref-bbox task aims to predict the bounding boxes corresponding to the described image regions.
The testB set of \citet{kazemzadeh2014referitgame}, the test-dev set of \citet{goyal2017making}, the test-dev-balanced set of \citet{hudson2019gqa}, and the test sets of other tasks are used for testing their corresponding tasks.
The weight of ShareGPT is set as $1.0$.

Furthermore, we present 7 zero-shot metrics employed by \citet{liu2024improved} to evaluate the generalization of methods: TextVQA \citep{singh2019towards}, POPE \citep{li2023evaluating}, MME \citep{fu2023mme}, ScienceQA (SQA) \citep{lu2022learn}, MMBench \citep{liu2023mmbench}, SEED-Bench-IMG ($\text{SEED}^\text{I}$) \citep{li2023seedbench}, and MM-Vet \citep{yu2023mmvet}.

\begin{table}[t!]
    \caption{Training data sizes and response format instructions for the fine-tuned tasks on the Academic Benchmark. The total training data size is 475k.}
    \vspace{-0.1in}
    \label{tab:appendix_dataset}
    \centering
    \scalebox{0.96}{
    \tabcolsep4pt
    {\renewcommand{\arraystretch}{1.1}
    \begin{tabular}{lcp{5.3cm}}
        \toprule[1.2pt]
        Tasks & Sizes & Response Format Instructions \\
        \midrule
        ShareGPT & 41k & -- \\
        ShareGPT4V & 98k & \\
        \midrule
        Ref-caption & 41k & Provide a short description for this region.\\
        \midrule
        VQAv2 & 83k & Answer the question using a single word \\
        GQA & 72k & or phrase.\\
        ChartQA & 18k & \\
        OCRVQA & 80k & \\
        \midrule
        Ref-bbox & 41k & Provide the bounding box coordinate of \\
         & &  the region this sentence describes. \\
        \bottomrule[1.2pt]
    \end{tabular}}}
    \vspace{-0.05in}
\end{table}

\begin{table}[t!]
    \caption{Response format instructions for the zero-shot test tasks on the Academic Benchmark.}
    \vspace{-0.1in}
    \label{tab:appendix_zero_dataset}
    \centering
    \scalebox{0.96}{
    \tabcolsep9pt
    {\renewcommand{\arraystretch}{1.1}
    \begin{tabular}{lp{5.3cm}}
        \toprule[1.2pt]
        Tasks & Response Format Instructions \\
        \midrule
        TextVQA & Answer the question using a single word\\
        POPE &  or phrase.\\
        MME & \\
        \midrule
        $\text{SQA}^\text{I}$ & Answer with the option's letter from the \\
        MMBench & given choices directly. \\
        $\text{SEED}^\text{I}$ & \\
        \midrule
        MM-Vet & --\\
        LLaVA-Bench & \\
        \bottomrule[1.2pt]
    \end{tabular}}}
    \vspace{-0.05in}
\end{table}

\begin{figure}[t!]
    \centering
    \includegraphics[width=0.98\linewidth]{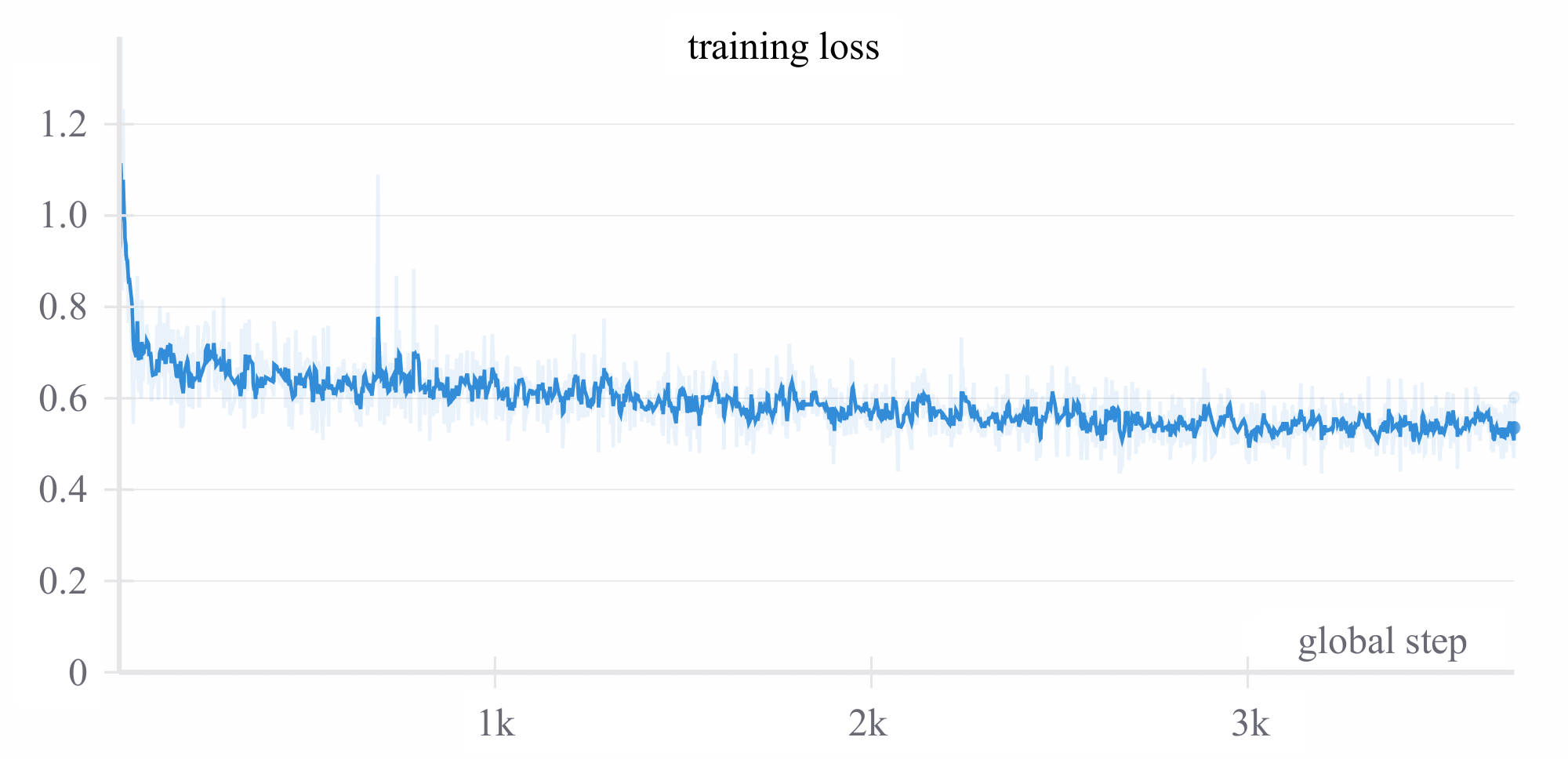}
    \vspace{-0.1in}
    \caption{Training loss of EW in the Academic Benchmark.}
    \label{fig:appendix_ew_loss}
    \vspace{-0.1in}
\end{figure}

\begin{table*}[t!]
   \caption{Complete results of the fine-tuned tasks on the Academic Benchmark using various pretrained models. $\bm{\Delta I\%}$ and $\bm{\Delta E\%}$ are the average per-task improvement and error on fine-tuned tasks compared to the STL baseline.}
   \vspace{-0.1in}
    \label{tab:appendix_pretrain_results}
    \centering
    \scalebox{0.96}{
    \tabcolsep6.4pt
    {\renewcommand{\arraystretch}{1.1}
    \begin{tabular}{l|l|ccccccc|cc}
        \toprule[1.2pt]
        \multirow{2}{*}{Models} & \multirow{2}{*}{Methods} & ShareGPT4V & Ref-caption & VQAv2 & GQA & ChartQA & OCRVQA & Ref-bbox & \multicolumn{2}{c}{\textbf{Overall}} \\
        \multirow{2}{*}{ } & \multirow{2}{*}{ } & CIDEr$\uparrow$ & CIDEr$\uparrow$ & EM$\uparrow$ & EM$\uparrow$ & EM$\uparrow$ & EM$\uparrow$ & IoU$\uparrow$ & \bm{$\Delta I$}\textbf{\%}\bm{$\uparrow$} & \bm{$\Delta E$}\textbf{\%}\bm{$\downarrow$} \\
        \midrule
        \multirow{3}{*}[-2.5pt]{LLaVA-v1.5-13B} & STL & 0.1407 & 0.5329 & 78.78 & 62.54 & 18.96 & 69.86 & 65.50 & & \\
        \cmidrule[0.5pt]{2-11}
        \multirow{3}{*}{} & EW & 0.1351 & 0.6008 & \textbf{79.51} & \textbf{63.05} & 22.28 & \textbf{69.41} & 68.44 & 4.55 & 0.66 \\
        \multirow{3}{*}{} & \cellcolor{lightblue}VisATB & \cellcolor{lightblue}\textbf{0.1391} & \cellcolor{lightblue}\textbf{0.6224} & \cellcolor{lightblue}79.33 & \cellcolor{lightblue}63.00 & \cellcolor{lightblue}\textbf{22.76} & \cellcolor{lightblue}69.22 & \cellcolor{lightblue}\textbf{73.35} & \cellcolor{lightblue}\textbf{6.89} & \cellcolor{lightblue}\textbf{0.29} \\
        \midrule
        \multirow{3}{*}[-2.5pt]{Qwen2-VL-2B} & STL & 0.1370 & 0.8402 & 81.11 & 64.55 & 64.36 & 73.40 & 27.89  & & \\
        \cmidrule[0.5pt]{2-11}
        \multirow{3}{*}{} & EW & 0.1305 & \textbf{0.6658} & 80.22 & \textbf{64.41} & 63.36 & \textbf{73.28} & \textbf{30.62} & -2.68 & 4.08  \\
        \multirow{3}{*}{} & \cellcolor{lightblue}VisATB & \cellcolor{lightblue}\textbf{0.1480} & \cellcolor{lightblue}0.6456 & \cellcolor{lightblue}\textbf{80.24} & \cellcolor{lightblue}64.23 & \cellcolor{lightblue}\textbf{64.04} & \cellcolor{lightblue}72.93 & \cellcolor{lightblue}30.46 & \cellcolor{lightblue}\textbf{-1.23} & \cellcolor{lightblue}\textbf{3.70} \\
        \bottomrule[1.2pt]
    \end{tabular}}}
    \vspace{-0.05in}
\end{table*}

\begin{table*}[t!]
    \caption{Complete results of the zero-shot tasks on the Academic Benchmark using various pretrained models. $\bm{\Delta I_{\text{zero}}\%}$ is the average per-task improvement in test performance on zero-shot tasks compared to the EW method.}
    \vspace{-0.1in}
    \label{tab:appendix_pretrain_zero_shot_results}
    \centering
    \scalebox{0.96}{
    \tabcolsep6.4pt
    {\renewcommand{\arraystretch}{1.1}
    \begin{tabular}{l|l|ccccccc|c}
        \toprule[1.2pt]
        Models & Methods & TextVQA$\uparrow$ & POPE$\uparrow$ & MME$\uparrow$ & SQA$\uparrow$ & MMBench$\uparrow$ & $\text{SEED}^\text{I}$$\uparrow$ & MM-Vet$\uparrow$ & \textbf{Overall (\bm{$\Delta I_{\textbf{zero}}$}\textbf{\%}\bm{$\uparrow$})} \\
        \midrule
        \multirow{2}{*}{LLaVA-v1.5-13B} & EW & \textbf{57.52} & \textbf{86.50} & 1556.48 & 64.18 & \textbf{58.25} & 64.56 & 35.40 & 0.00 \\
        \multirow{2}{*}{} & \cellcolor{lightblue}VisATB & \cellcolor{lightblue}57.06 & \cellcolor{lightblue}86.47 & \cellcolor{lightblue}\textbf{1567.57} & \cellcolor{lightblue}\textbf{65.90} & 57.65 & \cellcolor{lightblue}\cellcolor{lightblue}\textbf{64.58} & \cellcolor{lightblue}\textbf{35.80} & \cellcolor{lightblue}\textbf{0.38} \\
        \midrule
        \multirow{2}{*}{Qwen2-VL-2B} & EW & \textbf{73.45} & \textbf{88.41} & 1438.46 & 52.49 & \textbf{70.45} & \textbf{74.08} & \textbf{35.70} & 0.00 \\
        \multirow{2}{*}{} & \cellcolor{lightblue}VisATB & \cellcolor{lightblue}72.74 & \cellcolor{lightblue}\textbf{88.41} & \cellcolor{lightblue}\textbf{1445.53} & \cellcolor{lightblue}\textbf{64.35} & \cellcolor{lightblue}69.85 & \cellcolor{lightblue}73.69 & \cellcolor{lightblue}30.80  & \cellcolor{lightblue}\textbf{1.00} \\
        \bottomrule[1.2pt]
    \end{tabular}}}
    \vspace{-0.05in}
\end{table*}

The training data sizes and response format instructions for fine-tuned tasks are depicted in Table~\ref{tab:appendix_dataset}, while the response format instructions for zero-shot test tasks are presented in Table~\ref{tab:appendix_zero_dataset}.
Moreover, we employ multiple data processing and splitting strategies to reduce the computational cost and ensure the evaluation reliability:
\begin{enumerate}[leftmargin=16pt, topsep=0pt, itemsep=1pt, partopsep=0pt]
    \item For ShareGPT4V, data is randomly partitioned, with 2k allocated to a validation set, 2k to a test set, and the remainder reserved for training.
    \item For all VQA tasks and RefCOCO, training data from the same image are merged into a single conversation.
    \item For RefCOCO, training conversations are divided into segments, each containing fewer than 10 turns.
    \item For OCRVQA, 80k conversations are randomly sampled from the training set.
    \item For VQAv2, GQA and OCRVQA, 20k data are randomly sampled from the validation set.
    \item For ShareGPT, invalid conversations are filtered out following \citet{zheng2024judging}, and conversations that surpass 2048 tokens are truncated.
\end{enumerate}



\section{The Sampling Rate for Mini Subsets}
\label{sec:appendix_sampling_rate}
As discussed in Section~\ref{sec:itc}, mini subsets enable the model to understand the instruction demands of all tasks. 
Due to the remarkable few-shot learning capabilities of large models, incorporating a small amount of training data can significantly enhance the model's adherence to the task instructions.
In our experiments on the Academic Benchmark, the mini subset from each task is obtained by randomly sampling $1/32$nd of the entire dataset from that task.
As shown in Figure~\ref{fig:appendix_ew_loss}, this sampling ratio is informed by the decline pattern of training loss, which exhibits a rapid decrease during the initial 100+ steps, followed by a gradual reduction until reaching the final 3,700+ steps.
This pattern suggests that the model effectively grasps the instruction demands in the initial phase,  with the subsequent phase dedicated mainly to acquiring the knowledge embedded within the data.
Practically, we recommend ensuring that the mini subset from each task contains at least 1,000 data points, and the total number of training steps for all mini subsets exceeds 100.

\begin{table*}[t!]
    \caption{Detailed results of ablation studies for the fine-tuned tasks on the Academic Benchmark.}
    \vspace{-0.1in}
    \label{tab:appendix_full_ablation}
    \centering
    \scalebox{0.96}{
    \tabcolsep6pt
    {\renewcommand{\arraystretch}{1.1}
    \begin{tabular}{l|ccccccc|cc}
        \toprule[1.2pt]
        \multirow{2}{*}{Methods} & ShareGPT4V & Ref-caption & VQAv2 & GQA & ChartQA & OCRVQA & Ref-bbox & \multicolumn{2}{c}{\textbf{Overall}} \\
        \multirow{2}{*}{ } & CIDEr$\uparrow$ & CIDEr$\uparrow$ & EM$\uparrow$ & EM$\uparrow$ & EM$\uparrow$ & EM$\uparrow$ & IoU$\uparrow$ & \bm{$\Delta I$}\textbf{\%}\bm{$\uparrow$} & \bm{$\Delta E$}\textbf{\%}\bm{$\downarrow$} \\
        \midrule
        EW & 0.1411 & 0.5591 & 78.27 & 62.20 & 19.60 & 67.73 & 61.63 & 8.75 & 0.10 \\
        \midrule
        VisATB ($\bm{\alpha}\!=\![1,0,0]$) & 0.1448 & 0.5763 & 78.34 & 62.16 & 19.52 & 67.72 & 61.81 & \underline{9.67} & \underline{0.10} \\
        VisATB ($\bm{\alpha}\!=\![0,1,0]$) & 0.1333 & 0.5520 & 78.25 & 62.12 & 20.20 & 68.00 & 62.61 & 8.45 & \textbf{0.05} \\
        VisATB ($\bm{\alpha}\!=\![0,0,1]$) & 0.1455 & 0.5706 & 77.46 & 61.30 & 20.08 & 67.04 & 71.52 & \textbf{12.21} & 0.30 \\
        \midrule
        VisATB ($\bm{\alpha}\!=\![0.50,0.50,0]$) & 0.1340 & 0.5626 & 78.39 & 62.27 & 20.04 & 67.92 & 61.92 & 8.58 & \textbf{0.06} \\
        VisATB ($\bm{\alpha}\!=\![0.33,0.33,0.33]$) & 0.1321 & 0.5591 & 78.12 & 62.08 & 19.68 & 66.68 & 66.08 & \underline{8.77} & 0.32 \\
        \rowcolor{lightblue} VisATB ($\bm{\alpha}\!=\![0.25,0.25,0.50]$) & 0.1437 & 0.5724 & 77.99 & 61.81 & 20.16 & 67.48 & 67.38 & \textbf{11.29} & \underline{0.15} \\
        \midrule
        VisATB ($T\!=\!2.0$) & 0.1433 & 0.5642 & 78.10 & 62.08 & 20.16 & 67.67 & 63.06 & 9.92 & \underline{0.12} \\
        VisATB ($T\!=\!1.0$) & 0.1369 & 0.5752 & 78.15 & 62.09 & 20.32 & 67.71 & 65.00 & \underline{10.24} & \textbf{0.11} \\
        \rowcolor{lightblue} VisATB ($T\!=\!0.5$) & 0.1437 & 0.5724 & 77.99 & 61.81 & 20.16 & 67.48 & 67.38 & \textbf{11.29} & 0.15 \\
        \midrule
        VisATB (precise Diff) & 0.1345 & 0.5604 & 78.00 & 61.87 & 21.04 & 67.46 & 67.83 & \underline{10.75} & \underline{0.16} \\
        \rowcolor{lightblue} VisATB (real Diff) & 0.1437 & 0.5724 & 77.99 & 61.81 & 20.16 & 67.48 & 67.38 & \textbf{11.29} & \textbf{0.15} \\
        \bottomrule[1.2pt]
    \end{tabular}}}
    \vspace{-0.03in}
\end{table*}

\begin{table*}[t!]
    \caption{Detailed results of ablation studies for the zero-shot tasks on the Academic Benchmark.}
    \vspace{-0.1in}
    \label{tab:appendix_full_ablation_zero_shot_results}
    \centering
    \scalebox{0.96}{
    \tabcolsep6.1pt
    {\renewcommand{\arraystretch}{1.1}
    \begin{tabular}{l|ccccccc|c}
        \toprule[1.2pt]
        Methods & TextVQA$\uparrow$ & POPE$\uparrow$ & MME$\uparrow$ & SQA$\uparrow$ & MMBench$\uparrow$ & $\text{SEED}^\text{I}$$\uparrow$ & MM-Vet$\uparrow$ & \textbf{Overall (\bm{$\Delta I_{\textbf{zero}}$}\textbf{\%}\bm{$\uparrow$})} \\
        \midrule
        EW & 53.96 & 86.83 & 1524.41 & 60.74 & 48.54 & 54.13 & 29.00 & 0.00 \\
        \midrule
        VisATB ($\bm{\alpha}\!=\![1,0,0]$) & 54.40 & 86.87 & 1499.87 & 60.86 & 51.03 & 56.79 & 29.00 & \textbf{1.36} \\
        VisATB ($\bm{\alpha}\!=\![0,1,0]$) & 54.19 & 86.49 & 1511.93 & 59.59 & 39.09 & 48.14 & 30.60 & -3.96 \\
        VisATB ($\bm{\alpha}\!=\![0,0,1]$) & 53.87 & 86.68 & 1487.54 & 60.74 & 49.91 & 55.55 & 29.50 & \underline{0.63} \\
        \midrule
        VisATB ($\bm{\alpha}\!=\![0.50,0.50,0]$) & 54.73 & 86.53 & 1518.02 & 60.79 & 51.89 & 57.83 & 29.50 & \underline{2.32} \\
        VisATB ($\bm{\alpha}\!=\![0.33,0.33,0.33]$) & 54.07 & 86.47 & 1514.76 & 61.26 & 52.92 & 57.39 & 30.80 & \textbf{3.04} \\
        \rowcolor{lightblue} VisATB ($\bm{\alpha}\!=\![0.25,0.25,0.50]$) & 53.87 & 86.73 & 1501.86 & 61.07 & 51.46 & 58.05 & 29.30 & 1.87 \\
        \midrule
        VisATB ($T\!=\!2.0$) & 54.07 & 86.66 & 1498.19 & 61.02 & 49.66 & 54.89 & 29.30 & 0.50 \\
        VisATB ($T\!=\!1.0$) & 54.13 & 86.45 & 1499.72 & 60.29 & 49.48 & 56.21 & 29.10 & \underline{0.52} \\
        \rowcolor{lightblue} VisATB ($T\!=\!0.5$) & 53.87 & 86.73 & 1501.86 & 61.07 & 51.46 & 58.05 & 29.30 & \textbf{1.87} \\
        \midrule
        VisATB (precise Diff) & 54.25 & 86.77 & 1504.47 & 60.83 & 51.72 & 57.47 & 30.80 & \textbf{2.61} \\
        \rowcolor{lightblue} VisATB (real Diff) & 53.87 & 86.73 & 1501.86 & 61.07 & 51.46 & 58.05 & 29.30 & \underline{1.87} \\
        \bottomrule[1.2pt]
    \end{tabular}}}
    \vspace{-0.03in}
\end{table*}

\section{The Simpler Form of VisATB in the Chat Benchmark}
\label{sec:appendix_simple_visatb}
In the Chat Benchmark, there are no specific constraints on the output format.
Consequently, only the entire datasets of tasks are required, without the need for mini subsets, simplifying the form of VisATB.
Specifically, the inter-task contribution of Task $i$ to Task $j$ can be calculated as:
\begin{equation}
    C_{i\rightarrow j} = \frac{V_{j}(i) - V_{j}(\text{base})}{V_{j}(j) - V_{j}(\text{base})},
\end{equation}
where $V_{j}(i)$ represents the validation performance on Task $j$ of a model trained on Task $i$, $V_{j}(j)$ denotes the validation performance on Task $j$ of a model trained on Task $j$ itself, and $V_{j}(\text{base})$ signifies the validation performance on Task $j$ of a pretrained base model.
Additionally, the intra-task difficulty of Task $i$ can be computed as:
\begin{equation}
    D_{i} = 1 - \frac{V_{i}(\text{mini}_i)}{V_{i}(i)},
\end{equation}
where $V_{i}(\text{mini}_i)$ denotes the validation performance on Task $i$ of a model trained on the mini subset of Task $i$, and $V_{i}(i)$ signifies the validation performance on Task $i$ of a model trained on Task $i$ itself.


\section{Detailed Results and Settings on Various Pretrained Models}
\label{sec:appendix_pretrain}

The detailed results on the Academic Benchmark using various pretrained models are shown in Tables~\ref{tab:appendix_pretrain_results} and \ref{tab:appendix_pretrain_zero_shot_results}.
The temperature $T$ is set as $0.5$ on LLaVA-v1.5-13B and $1.0$ on Qwen2-VL-2B.

\section{Detailed Ablation Results}
\label{sec:appendix_full_ablation}
The detailed ablation results on the Academic Benchmark are presented in Tables~\ref{tab:appendix_full_ablation} and \ref{tab:appendix_full_ablation_zero_shot_results}.

\section{The Precise Calculation Approach for Intra-Task Difficulty}
\label{sec:appendix_precise}
In the precise calculation approach for intra-task difficulty, the intra-task difficulty of Task $i$ can be calculated as follows:
\begin{equation}
    D_{i} = 1 - \frac{V_{i}(\text{mini}_i)}{V_{i}(i)},
\end{equation}
where $V_{i}(\text{mini}_i)$ denotes the validation performance on Task $i$ of a model trained on the mini subset of Task $i$, and $V_{i}(i)$ signfies the validation performance on Task $i$ of a model trained on Task $i$ itself.

\section{The Sampling Rate for Sufficiently Large Subsets}
\label{sec:appendix_sampling_rate_large}
As validated by experimental results in Sections~\ref{sec:m3it} and \ref{sec:academic}, $1/4$th subsets in the M$^3$IT Benchmark and the entire datasets in the Academic benchmark are sufficient for VisATB to accurately measure inter-task contribution and intra-task difficulty. 
Specifically, the $1/4$th subset of VQA in the M$^3$IT Benchmark contains 14k samples, while the entire dataset of ChartQA in the Academic benchmark comprises 18k samples.
Therefore, we recommend that subsets containing more than 10k samples and trained for over 100 steps are sufficiently large to ensure effective training of VisATB.

\end{document}